\documentclass{article}

\usepackage{arxiv}

\usepackage[utf8]{inputenc} %
\usepackage[T1]{fontenc}    %
\usepackage{hyperref}       %
\usepackage{url}            %
\usepackage{booktabs}       %
\usepackage{amsfonts}       %
\usepackage{nicefrac}       %
\usepackage{microtype}      %
\usepackage{lipsum}
\usepackage{graphicx}
\usepackage{easyReview}

\usepackage{caption}
\usepackage{subcaption}
\usepackage{amsmath}
\usepackage{amsfonts}
\usepackage{amsthm}
\usepackage{amssymb}
\usepackage{bbm}
\usepackage{bm}
\usepackage{lipsum}
\usepackage[numbers]{natbib}
\usepackage{stackengine} %
\usepackage{tabularx}
\usepackage{makecell}
\usepackage{booktabs}
\usepackage{diagbox}
\usepackage{enumitem}

\usepackage{thmtools, thm-restate}
\newtheorem{theorem}{Theorem}[section]
\newtheorem{lemma}[theorem]{Lemma}

\newtheorem*{proposition*}{Proposition}

\newtheorem{example}[theorem]{Example}

\usepackage{hyperref}
\hypersetup{
	colorlinks=true,
	linkcolor=blue,
	filecolor=magenta,
	urlcolor=cyan,
}

\usepackage{mathtools}

\usepackage{algorithm}
\usepackage{algpseudocode}

\usepackage{optidef}

\usepackage{xfrac}

\usepackage{multirow}%
\usepackage{hhline}%

\usetikzlibrary{decorations.pathreplacing,calligraphy,calc} %
\usepackage{xcolor}
\definecolor{darkgreen}{RGB}{9, 133, 9}

\DeclareMathOperator{\Tr}{Tr}
\newcommand{\norm}[1]{\left\lVert#1\right\rVert}

\newcommand{\matr}[1]{\bm{#1}}     %

\DeclareMathOperator{\fro}{F}

\usepackage{xspace}
\newcommand{\nlevels}{n_\text{levels}\xspace}
\newcommand{\mya}{\sum_{j=1}^n\widetilde{\lambda}_j}
\newcommand{\myb}{\sum_{i=1}^N\widetilde{\lambda}_i}

\date{}

\title{Deep Kernel Principal Component Analysis for Multi-level Feature Learning}
\date{}

\author{Francesco Tonin\thanks{Corresponding authors} , Qinghua Tao$^\ast$, Panagiotis Patrinos and Johan A.K. Suykens\\
Department of Electrical Engineering, ESAT-STADIUS,\\
KU Leuven. Kasteelpark Arenberg 10, B-3001 Leuven, Belgium\\
\texttt{\{francesco.tonin,qinghua.tao,panos.patrinos,johan.suykens\}@esat.kuleuven.be} \\
}

\begin{document}
\maketitle
\begin{abstract}
        Principal Component Analysis (PCA) and its nonlinear extension Kernel PCA (KPCA) are widely used across science and industry for data analysis and dimensionality reduction. %
        Modern deep learning tools have achieved great empirical success, but a framework for deep principal component analysis is still lacking.
        Here we develop a deep kernel PCA methodology (DKPCA) to extract multiple levels of the most informative components of the data.
        Our scheme can effectively identify new hierarchical variables, called deep principal components, capturing the main characteristics of high-dimensional data through a simple and interpretable numerical optimization.
        We couple the principal components of multiple KPCA levels, theoretically showing that DKPCA creates both forward and backward dependency across levels, which has not been explored in kernel methods and  yet is crucial to extract more informative features.
        Various experimental evaluations on multiple data types show that DKPCA finds more efficient and disentangled representations with higher explained variance in fewer principal components, 
    compared to the shallow KPCA. We demonstrate
	that our method allows for effective hierarchical data exploration, with the ability to separate the key generative factors of the input data both for large datasets and when few training samples are available.
        Overall, DKPCA can facilitate the extraction of useful patterns from high-dimensional data by learning more informative features organized in different levels, giving diversified aspects to explore the variation factors in the data, while maintaining a simple mathematical formulation.
\end{abstract}

\section{Introduction}

Principal Component Analysis (PCA) is a popular technique for dimensionality reduction~\cite{pca} and has been widely used in many fields~\cite{lever2017points}. In fact, high-dimensional data  are very common in data science when multiple variables are used to describe one sample; e.g., in biology, PCA has been applied to mass spectrometry, where thousands of proteins can be quantitatively profiled~\cite{ringner2008principal}. PCA  learns the most effective principal components to successfully reduce  the dimensionality of the data while retaining most of the trends and patterns. This
 relies on the assumption that the given observations lie in a lower-dimensional linear subspace. Under this assumption, PCA seeks the best low-rank representation of the given data. PCA can be efficiently computed using the Singular Value Decomposition (SVD) and is optimal when  data are corrupted by small Gaussian noises \cite{wright2009}.
 Real-world data commonly show nonlinear relationships, so, for nonlinear problems, PCA can be extended to Kernel PCA (KPCA)~\cite{scholkopf1998}, which  manages to simplify  such complexity and high dimensionality  to extract useful patterns in nonlinear subspaces. KPCA first maps the inputs to a high-dimensional feature space and then applies PCA to the mapped features  either through nonlinear feature mappings in the
 primal or equivalently kernel functions in the dual. 
 In the Lagrange dual formulation of KPCA, the feature map does not need to be explicitly defined and positive-definite kernel functions are instead used by Mercer's theorem \cite{mercer1909}.

In deep learning, dimensionality reduction and learning informative features are also widely studied through the latent space models, such as Variational Autoencoders (VAEs) \cite{vae}, which have become popular tools to extract  latent 
	features describing the factors of variation in the given training distribution. These models assume that there exists a prior distribution $p(\bm z)$ over a small number of ground-truth factors of variation, such that an observation $\bm x$ is obtained by first sampling $\bm z$ from $p(\bm z)$ and then sampling from a conditional distribution $p(\bm x|\bm z)$. In this setting, the goal is to find a representation of the data that learns the factors of variation in $\bm z$ independently, i.e., that disentangles the factors of variation. State-of-the-art models for disentangled feature learning include InfoGAN \cite{infogan}, Restricted Boltzmann machines \cite{reed2014,hinton2006}, $\beta$-VAE \cite{betavae} and its variants  \cite{factorvae,mig}. For instance, in $\beta$-VAE, $p(\bm z) = \mathcal{N}(0,\matr{I})$ and the encoder $q(\bm z|\bm x)$ is matched to the prior $p(\bm z)$ by minimizing the Kullback–Leibler divergence $D_\text{KL}(q(\bm z|\bm x) || p(\bm z))$. Neural networks are used to model the generative model with probabilistic encoder $q(\bm z|\bm x)$ and decoder $p(\bm x|\bm z)$ \cite{vae}. A recent large-scale extensive experimental research has shown that the performance of VAE-based models varies greatly with random initialization, hyperparameters, and dataset, so 
	reliable extraction of independent components describing the variation factors of data remains challenging \cite{locatello}.

While (K)PCA has been widely used in science and industry, the modelling flexibility of using a single feature mapping or kernel function can be insufficient and it also cannot learn well-disentangled representations \cite{betavae}. 
For such feature-learning tools, disentanglement of the variation factors (components) in the data is highly desirable \cite{bengio2009,bengio2013representation} and it has been suggested that disentangled representations can benefit interpretation analysis, e.g., in the medical domain \cite{holzinger2019causability,sarhan2019learning}. 
For instance, a model trained on gene expression data may learn components such as the cell type or the cell state. 
In addition, because (K)PCA is a shallow model employing a single feature mapping, it learns only one flat level of components.
On the other hand, deep learning has achieved pervasive empirical success with great modelling flexibility \cite{goodfellow2016deep}, but a framework combining deep architectures and principal component analysis remains lacking.

Deep kernel learning 
tackles multiple latent spaces for greater flexibility, more informative
hierarchical investigation of the data, and  kernel-based  interpretations.
	There exist many works in deep kernel learning considering supervised learning (see \cite{bohn2019} and references therein), but little investigation has been spared  on the unsupervised settings, though a concatenation of operator-valued kernel layers was considered  for data autoencoding in  \cite{laforgue2019}.	
In \cite{deng2019}, it is proposed to conduct the shallow PCA  to extract principal components, which are then  applied to another KPCA, where each KPCA independently and sequentially optimizes its variance maximization. 
 Importantly, %
	when extending %
to deep	architectures, \cite{bfc} warn that simply doing a sequential kernel learning is not enough to achieve good accuracy  due to the lack of	%
backward feature correction, meaning that shallow	%
layers need to use %
the information from deeper layers	to boost their own learned representation. In \cite{bfc}, it is proved %
that hierarchical learning cannot be efficiently achieved without backward feature correction.

\begin{figure}[t]
    \centering
	\includegraphics[width=0.6\textwidth]{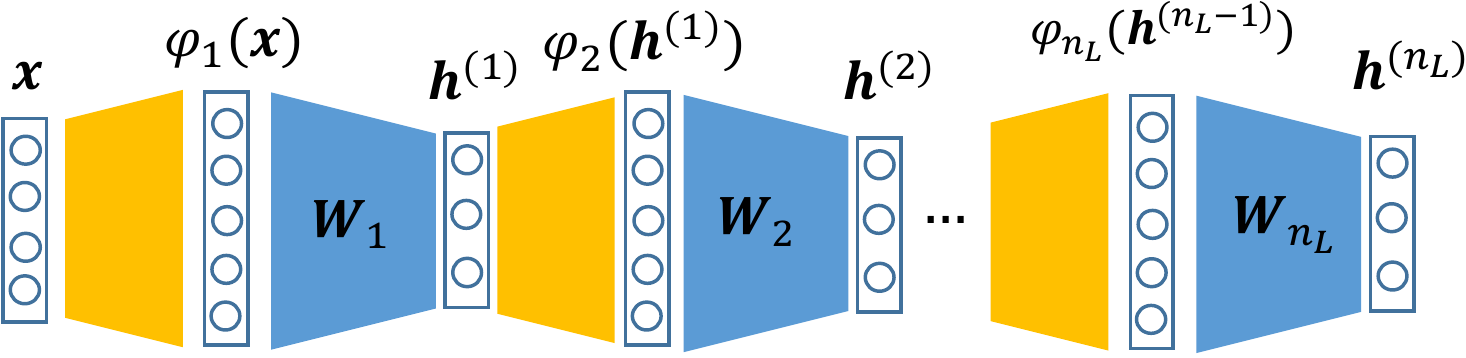}
    \caption{Topology of the RKM-based deep KPCA with $n_L$ levels. An input vector $\matr{x}$ is mapped to the feature space of the first level using a feature map $\varphi_1$ with hidden features $\matr{h}^{(1)}$ in the latent space of the first level. Subsequently, the input of level $j$, with feature map $\varphi_j$, are the hidden features of level $j-1$.}
    \label{fig:topology}
\end{figure}

{
	In this paper, we establish a novel Deep Kernel Principal Component Analysis (DKPCA) framework with the following main aspects.
	\begin{itemize}
	    \item DKPCA presents multiple levels of principal components associated with the key properties of the data for more
informative %
feature learning in multiple subspaces.
    The	objective of each level is attained as an upper bound of a shallow KPCA\footnote{To differentiate  DKPCA, we name the classical KPCA as shallow KPCA considering its one-level architecture.} problem, and multiple levels are %
constructed by coupling the latent space of level $j-1 (j\geq 2)$ with the input space of level $j$, where the depth is given by the learned spaces directly relating to the principal components, as shown in Fig. \ref{fig:topology}.
 We derive that the optimization problem of our method explicitly formulates
a set of nonlinear equations for each level resembling an eigenvalue problem of some matrix $\matr{M}_j$, in contrast with black-box optimization in deep learning.
	\item Interestingly, %
	$\matr{M}_j$ %
fuses the	hidden features of previous and subsequent levels. %
This means that the proposed deep architecture introduces not only	\textit{forward} couplings between the levels, but also \textit{backward} couplings, which by far has not been explored in kernel methods and yet is crucial for effective hierarchical representation learning according to the theoretical analysis in \cite{bfc}. 
As the levels are coupled together, we formulate a multi-level constrained optimization problem with an eigenvalue problem at each level with hidden features as optimization variables, also facilitating deep approximation analysis of the given data.
\item The solution of the proposed optimization process	%
	gives both the deep eigenvectors and the deep eigenvalues of the DKPCA: they correspond to the solution of the eigenvalue problem of each level. Within the considered deep architecture, we then construct a generative procedure for the DKPCA by defining both an out-of-sample encoding scheme and a decoding procedure, discussing connections with Autoencoders. %
	The generative procedure generates 
	new samples from  multiple latent
	spaces in different levels, makes it possible to explore the role of the deep eigenvectors of each level through the latent space traversals, and gives diversified aspects to explore the variation factors of  data. Our method can also be implemented with out-of-sample extensions which allow to efficiently tackle large-scale cases.
\end{itemize}
}

Extensive numerical experiments demonstrate the efficacy and advantages of the proposed DKPCA from
different aspects and in different tasks on multiple data types. 1) DKPCA gives higher explained variance
than shallow KPCA, indicating that more information is captured in fewer components. We also provide a
strategy for practitioners to select the numbers of components and levels, which is in contrast with typical
deep learning tools that use trial and error strategies in determining the network structure. 2) DKPCA
effectively facilitates hierarchical data exploration, as the role of each principal component in each level
can be investigated through the generation of new data. In images of 3D objects with different generative
factors (i.e., colors, size, etc.), our deep method creates a learning hierarchy in the components in each
level. Prevailing features are typically learned in the shallower levels, e.g., colors, while the deeper levels
capture more subtle features, e.g., the specific object shape. 3) Quantitative performances are evaluated by
comparing to state-of-the-art methods in disentangled feature learning \cite{betavae,mig,factorvae} when few training samples are available, which is of particular interest in many
real-world problems where data are difficult or expensive to collect. 4) We show that the more informative
features extraction by DKPCA can be applied to multiple data types benefiting various downstream tasks in
data science, such as regression and classification.

\section{Background} \label{sec:background}
This section describes the shallow KPCA problem introducing its formulation in the RKM framework through the Fenchel-Young inequality. 
The RKM formulation of KPCA %
gives another expression of the
Least-Squares Support Vector Machine (LS-SVM) KPCA problem
\cite{suykens2003} with visible and hidden units similar to
the energy of Restricted Boltzmann Machines (RBMs)
\cite{bengio2009,fischer2014,hinton2006,salakhutdinov2015}. In this new
formulation, contrary to RBMs, both the visible units  and the hidden
units can be continuous. To derive this formulation, consider 
training data $D=\{\bm x_i\}^N_{i=1}$ with $\bm x_i \in \mathbb{R}^d$, a feature
map $\varphi: \mathbb{R}^d \mapsto \mathbb{R}^{d_{\mathcal{F}}}$, and let
$s$ be the number of selected principal components. In the LS-SVM setting,
the KPCA problem can be written as minimizing a regularization term
and finding directions of maximum variance \cite{suykens2002}:
\begin{mini}|l|
	{\matr{W},\bm e_i}{J_{\text{kpca}} = \frac{\eta}{2}\Tr{(\matr{W}^\top \matr{W})}-\frac{1}{2}\sum_{i=1}^N\boldsymbol e_i^\top \matr{\Lambda}^{-1} \boldsymbol e_i}{}{}
	\label{eq:kpca}
	\addConstraint{\boldsymbol e_i}{= \matr{W}^\top \varphi(\boldsymbol x_i),}{\quad i=1,\dots,N,}
\end{mini}
where $\matr{W} \in \mathbb{R}^{d_{\mathcal{F}} \times s}$ is the interconnection matrix, $\boldsymbol e_i \in \mathbb{R}^{s}$ are the score variables along the selected $s$ projection directions, and $\matr{\Lambda}={\rm diag}\{\lambda_1, \ldots, \lambda_s\} \succ 0, \eta > 0$ are regularization hyperparameters. 

The RKM formulation of KPCA \cite{drkm} is given by an upper bound of $J_{\text{kpca}}$ obtained component-wise with the Fenchel-Young inequality 
$\frac{1}{2\lambda}e^2+\frac{\lambda}{2}h^2 \geq eh, \, \forall e, h \in \mathbb{R}$
which introduces the hidden features $\bm h$  and leads to the following objective with conjugate feature duality:

\begin{equation} \label{eq:rkm-kpca}
	\overline{J}_{\rm kpca} = -\sum_{i=1}^N \varphi(\bm  x_i)^\top  \matr{W} \bm h_i+\frac{1}{2}\sum_{i=1}^N {\bm h_i}^\top\matr{\Lambda}  \bm h_i+\frac{\eta}{2}\Tr{\left( \matr{W}^\top \matr{W} \right)},
\end{equation}
where $\bm h_i\in \mathbb R^s$ are the conjugated hidden features corresponding to each training sample $\bm x_i$; in representation learning, $\bm h_i$ is also known as the latent representation of $\bm x_i$ consisting of $s$ latent variables or of $s$ hidden features. 
Note that the first term of \eqref{eq:rkm-kpca} is similar to the energy of an RBM with connections between visible units $\matr{x}_i$ in the input space and hidden units $\matr{h}_i$ in the latent space. 
The stationary point conditions of $\overline{J}_{\rm kpca}(\matr{W},\bm h_i)$ are given by:
\begin{equation} \label{eq:kpca:stationarity}
\begin{cases}
\dfrac{\partial \overline{J}_{\rm kpca}(\matr{W},\bm h_i)}{\partial \bm h_i}=0 \implies \matr{W}^T \varphi(\bm x_i) = \matr{\Lambda} \bm h_i, \, \forall i=1,\dots,N \\
\dfrac{\partial \overline{J}_{\rm kpca}(\matr{W},
\bm h_i)}{\partial \matr{W}}=0 \implies \matr{W} = \dfrac{1}{\eta} \sum\limits_{i=1}^N \varphi(\bm x_i) \bm h_i^T.
\end{cases}
\end{equation}
Eliminating $\matr{W}$ and considering a positive definite kernel function $k: \mathbb{R}^d \times \mathbb{R}^d \mapsto \mathbb{R}$ with $k(\bm x_i,\bm x_j)=\varphi(\bm x_i)^\top \varphi(\bm x_j)$, the stationary points of $\overline{J}_{\rm kpca}(\matr{W},\bm h_i)$ 
are given in the dual by the following eigenvalue problem 

\begin{equation}\label{eq:eigen:kpca}
	\frac{1}{\eta} \matr{K} \matr{H} = \matr{H} \matr{\Lambda},
\end{equation}
where $\matr{K}\in \mathbb R^{N\times N}$
denotes the kernel matrix induced by $k(\cdot, \cdot)$ and  the matrix $\matr{H} = [\bm h_1, \dots, \bm h_N]^\top$ incorporates the conjugate hidden features for all $N$ data points. 
In  \eqref{eq:eigen:kpca}, the  hidden features $\matr{H}$ conjugated along $s$ projection directions now correspond to the first $s$ eigenvectors, with the first $s$ eigenvalues corresponding to the hyperparameter $\matr{\Lambda}$ in \eqref{eq:rkm-kpca}. Meanwhile, $\eta$ becomes a scaling coefficient that does not change the solution space, and thus can be simply set as $1$.
Note that, in the conjugate feature duality of RKMs, the dual variables $\matr{h}$ correspond to the latent variables playing the role of hidden features living in the latent space.

The dual problem \eqref{eq:eigen:kpca} corresponds to the kernel PCA problem as defined in \cite{scholkopf1998}. While \eqref{eq:eigen:kpca}  is regularized by normalizing the eigenvectors to the unit ball in feature space, the primal problem \eqref{eq:rkm-kpca} is explicitly regularized with coefficients $\matr{\Lambda},\eta$ chosen at the hyperparameter selection level.
Each eigenvalue/eigenvector pair 
corresponds to a principal component in KPCA. Therefore, for the first $s$ principal components, one can solve the dual problem \eqref{eq:eigen:kpca} by considering the $s$ largest eigenvalues and their eigenvectors, which lead to  $\overline{J}_{\rm kpca}=0$.
Since $\overline{J}_{\rm kpca}$ is unbounded below regarding its optimization in the primal, \cite{drkm} proposed to instead minimize a stabilized version to make the objective suitable for minimization, such that
$
	\overline{J}_{\rm kpca, stab} = \overline{J}_{\rm kpca} + \frac{c_{\text{stab}}}{2} \overline{J}_{\rm kpca}^2,
$
where $c_{\text{stab}} > 0$ is a hyperparameter. It can be shown that $\overline{J}_{\rm kpca}$ and $\overline{J}_{\rm kpca, stab}$ share the same stationary points \cite{pandey2021}.

Deep kernel methods based on the RKM framework were considered in \cite{drkm,tonin2021}.
In \cite{drkm}, the KPCA levels are used as feature extractors for regression and classification. For these supervised learning tasks, \cite{drkm} described a heuristic algorithm for the case of linear kernels  with a level-wise forward phase, while the backward phase is only considered  from the last level to the first one, 
discarding backward connections of all intermediate levels.
Furthermore, \cite{drkm} did not deal with the interpretation of the induced eigenvalues
in the deep RKM; in this work, we detail the role of different eigenvalues
in relation to the importance of each level and its principal components.
In \cite{tonin2021}, a two-level architecture for unsupervised learning was considered with orthogonality constraints on the latent variables 
within each level and between the levels. 
By formulating the constraints into a penalty term in the objective,
a straightforward numerical approach was employed to solve such unconstrained optimization problem, 
where the backward couplings between 
the hidden features
were omitted. 
Though cast in the RKM framework, in this paper we consider  more general deep KPCA architectures with multiple levels and latent spaces  through the lens of  a set of level-wise shallow KPCA problems, and importantly both forward and backward dependencies
between levels are involved. Thanks to such new 
problem formulation, novel training schemes are proposed together with theoretical error bounds, where a generative model and the out-of-sample extension are also discussed, demonstrating  empirical evidence of the advantages of our deep architectures and facilitating interpretations of the obtained deep principal components.

Another form of deep KPCA was proposed in \cite{deng2019}, where PCA was firstly conducted to extract principal components of the data and then further dimensionality reduction was sequentially applied to the extracted features from the previous (K)PCA layer.
This serial approach makes
each layer straightforwardly optimize its variance maximization objective, which is independent of  other layers.

\section{Deep Kernel Principal Component Analysis} \label{sec:dkpca}
In this section, we present the proposed DKPCA. We start by describing the model formulation of DKPCA. Next, we derive the optimization algorithm. Finally, the generative DKPCA model is introduced.

\subsection{DKPCA Model Formulation}

We construct the objective function of DKPCA by joining the KPCA objectives of multiple levels in the Restricted Kernel Machine (RKM) framework \cite{drkm}, which
combines the flexibility of deep architectures 
	and the interpretations rooted in kernel methods. 
DKPCA considers general cases consisting of $\nlevels$ ($\nlevels \geq 2$) KPCA levels
stacked in the corresponding latent spaces, i.e., the hidden features of level $j$ are the input of level $j+1$, inducing inter-level couplings, similar to the stacked Autoencoders  \cite{bengio2009}. 
Correspondingly, 
the objective for the proposed DKPCA is formulated in the primal model representation:
\begin{align} \label{eq:deeprkm:primal:main}
	\begin{split}
		J = &-\sum_{i=1}^N \varphi_1(\bm x_i)^\top  \matr{W}_1 \bm h_i^{(1)}+\frac{1}{2}\sum_{i=1}^N {\bm h_i^{(1)}}^\top \matr{\Lambda}_1 \bm h_i^{(1)}+\frac{\eta_1}{2}\Tr{\left( \matr{W}_1^\top \matr{W}_1 \right)}\\
		&+\sum_{j=2}^{\nlevels} \left[ -\sum_{i=1}^N \varphi_j(\bm h_i^{(j-1)})^\top  \matr{W}_j \bm h_i^{(j)}+\frac{1}{2}\sum_{i=1}^N {\bm h_i^{(j)}}^\top \matr{\Lambda}_j \bm h_i^{(j)}+\frac{\eta_j}{2}\Tr{\left( \matr{W}_j^\top \matr{W}_j \right)} \right].
	\end{split}
\end{align}
The feature map $\varphi_1: \mathbb{R}^d \mapsto \mathbb{R}^{d_{\mathcal{F}_1}}$ of the first level takes the original data as the input, while $\varphi_j: \mathbb{R}^{s_j} \mapsto \mathbb{R}^{d_{\mathcal{F}_j}}$ is the feature map of level $j=2,\dots,\nlevels$ that  takes the hidden features $\bm h_i^{(j-1)}$ of level $j-1$ as the input, where $\matr{W}_j \in \mathbb{R}^{d_{\mathcal{F}_j} \times s_j}$ is the interconnection matrix of level $j$. Here, the matrix $\matr{H}_j = [\bm h_1^{(j)}, \dots, \bm h_N^{(j)}]^\top \in \mathbb R^{N \times s_j}$ incorporates the hidden features conjugated along $s_j$  projection directions for all $N$ data points, where $s_j$ is the number of selected principal components by the $j$-th level of our DKPCA. In the primal formulation, $\matr{\Lambda}_j={\rm diag}\{\lambda^{(j)}_1, \ldots, \lambda^{(j)}_{s_j}\}$ and $\eta_j \neq 0$ both serve as the hyperparameters of level $j$. 
While $\eta>0$ in the shallow KPCA case for variance maximization in \eqref{eq:kpca}, this constraint is not required in the deep objective \eqref{eq:deeprkm:primal:main}, having complex inter-level couplings.
Note that, in our DKPCA formulation, the visible units $\bm x_i$ in the input space are conjugated with the multi-level hidden features $\bm h_i^{(j)}$ in the  latent space of each level $j$,
	giving an energy function that resembles the deep Boltzmann machine \cite{salakhutdinov2009}.
The DKPCA topology in its primal formulation is visualized in Fig. \ref{fig:topology}.

{
The projection directions of shallow (K)PCA are uncorrelated due to the orthogonality of different principal components as in \eqref{eq:eigen:kpca}. Similarly for DKPCA,  we impose intra-level orthogonality on $\matr{H}_j$, i.e., $\matr{H}_j^\top \matr{H}_j=\matr{I}$.
From the stationary points of \eqref{eq:deeprkm:primal:main}, the formulation of DKPCA in the dual variables is:
\begin{equation}
	\label{eq:system:nl:main}
	\left\{
	\begin{array}{lll}
		\text{Level 1: }          & \left[ \dfrac{1}{\eta_1} \matr{K}_1 + \dfrac{1}{\eta_2} {\matr{\mathcal{G}}_1(\matr{H}_1,\matr{H}_2)} \matr{H}_1^\top \right] \matr{H}_1                  & = \matr{H}_1 \matr{\Lambda}_1, \vspace{3mm}      \\
		
		\text{Level $j$: }        & \left[ \dfrac{1}{\eta_j} \matr{K}_j(\matr{H}_{j-1}) + \dfrac{1}{\eta_{j+1}} {\matr{\mathcal{G}}_j(\matr{H}_j,\matr{H}_{j+1})} \matr{H}_j^\top \right] \matr{H}_j & = \matr{H}_j \matr{\Lambda}_j, \  \forall j=2,\dots,\nlevels-1,
		\vspace{3mm}      \\
		
		\text{Level $\nlevels$: } & \dfrac{1}{\eta_{\nlevels}} \matr{K}_{\nlevels}(\matr{H}_{\nlevels-1}) \; \matr{H}_{\nlevels}                                         & = \matr{H}_{\nlevels} \matr{\Lambda}_{\nlevels}.
	\end{array}
	\right.
\end{equation}
\tikzstyle{startstop} = [rectangle, rounded corners, minimum width=3cm, minimum height=1cm,text centered, draw=black, fill=orange!30]
\tikzstyle{arrow} = [thick,->,>=stealth]
\begin{figure}[t]
\centering
\begin{tikzpicture}[]
\node (level1) [startstop] {\makecell[c]{Level 1\vspace{0.15cm}\\$\matr{K}_1(\matr{X})$\\$+ {\matr{\mathcal{G}}_1(\matr{H}_1,\matr{H}_2)} \matr{H}_1^\top$}};
\node (level2) [startstop,right of=level1,xshift=4.75cm] {\makecell[c]{Level 2\vspace{0.15cm}\\$\matr{K}_2(\matr{H}_{1})$\\$+{\matr{\mathcal{G}}_2(\matr{H}_2,\matr{H}_{3})} \matr{H}_2^\top$}};
\node (leveln) [startstop,right of=level2,xshift=6cm] {\makecell[c]{Level $n_L$\vspace{0.25cm}\\$ \matr{K}_{n_L}(\matr{H}_{n_L-1})$\vspace{0.25cm}}};
\draw [arrow,color=darkgreen] ([yshift=0.3cm]level1.east) -- node[anchor=south] {$\matr{H}_1$} ([yshift=0.3cm]level2.west);
\draw [arrow,color=blue] ([yshift=-0.3cm]level2.west) -- node[anchor=north] {$\matr{H}_2$} ([yshift=-0.3cm]level1.east);
\node (dots) at ($(level2)!.5!(leveln)$) {\ldots};
\draw [arrow,color=darkgreen] ([yshift=0.3cm]level2.east) -- node[anchor=south] {$\matr{H}_2$} ([yshift=0.3cm,xshift=-0.2cm]dots.west);
\draw [arrow,color=blue] ([yshift=-0.3cm,xshift=-0.2cm]dots.west) -- node[anchor=north] {$\matr{H}_3$} ([yshift=-0.3cm]level2.east);
\draw [arrow,color=darkgreen] ([yshift=0.3cm,xshift=0.2cm]dots.east) -- node[anchor=south] {$\matr{H}_{n_L-1}$} ([yshift=0.3cm]leveln.west);
\draw [arrow,color=blue] ([yshift=-0.3cm]leveln.west) -- node[anchor=north] {$\matr{H}_{n_L}$} ([yshift=-0.3cm,xshift=0.2cm]dots.east);
\draw [arrow,color=darkgreen] ([yshift=0.5cm]level1.north west) -- node[anchor=south] {Forward Couplings} ([yshift=0.5cm]leveln.north east);
\draw [arrow,color=blue] ([yshift=-0.5cm]leveln.south east) -- node[anchor=north] {Backward Couplings} ([yshift=-0.5cm]level1.south west);
\end{tikzpicture}
\caption{Graphical illustration of the DKPCA dual problem \eqref{eq:system:nl:main} with $n_L$ levels. Each arrow  goes from the level that is characterized by the corresponding hidden features to the level where it is used as input. DKPCA introduces not only \textit{forward} couplings (green arrows), but also \textit{backward} couplings (blue arrows) between the levels. For simplicity, $\eta_j=1$ in the diagram.}
\label{fig:dual}
\end{figure}
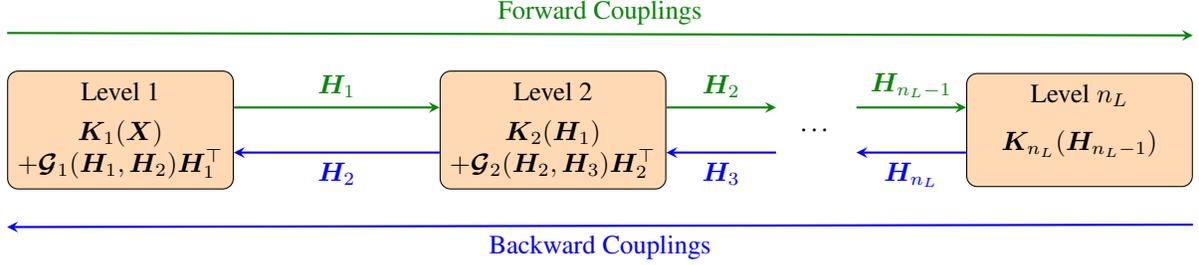
A graphical illustration of \eqref{eq:system:nl:main} is given in Fig. \ref{fig:dual}. The kernel matrices are obtained as follows: $\matr{K}_1 \in \mathbb{R}^{N \times N}$ is attained as $(K_1)_{ik} = k_1(\bm x_i,\bm x_k)$ and $\matr{K}_j \in \mathbb{R}^{N \times N}$ as  $(K_j)_{ik} = k_j(\bm  h_i^{(j-1)},\bm h_k^{(j-1)})$, where  $k_1: \mathbb{R}^d \times \mathbb{R}^d \mapsto \mathbb{R}$ with $k_1(\bm x, \bm y) = \varphi_1(\bm x)^\top \varphi_1(\bm y)$ is the kernel function of the first level and $k_j: \mathbb{R}^{s_{j-1}} \times \mathbb{R}^{s_{j-1}} \mapsto \mathbb{R}$  with $k_j(\bm x, \bm y) = \varphi_j(\bm x)^\top \varphi_j(\bm y)$ is the kernel function of level $j=2,\dots,\nlevels$  by the kernel trick. Instead of first defining a feature map $\varphi_j$, one can simply choose a positive definite kernel $k_j$ due to Mercer’s theorem \cite{mercer1909}, guaranteeing the existence of a feature map $\varphi_j$ such that $k_j(y,z) = \varphi_j(y)^\top \varphi_j(z)$.
The stationary points of \eqref{eq:deeprkm:primal:main} can be found in \ref{sec:methods:dkpca}.

In \eqref{eq:system:nl:main}, $\matr{\mathcal{G}}_j(\matr{H}_j,\matr{H}_{j+1}) \in \mathbb{R}^{N \times s_j}$ are matrices jointly depending on the conjugated hidden features $\matr{H}_j$ and $\matr{H}_{j+1}$.
 In particular, they are formulated as  $\matr{\mathcal{G}}_j(\matr{H}_j,\matr{H}_{j+1}) = \left( \matr{I}_N \odot  \left(\matr{H}_{j+1}\matr{H}_{j+1}^\top\right) \right)^\top \matr{J}_{K_{j+1}}(\matr{H}_j) \in \mathbb{R}^{N \times s_j}$,
 where the Khatri-Rao product between matrices $\matr{A} = \left[ \bm a_1 \; \cdots \; \bm a_n \right] \in \mathbb{R}^{m_1 \times n}$ and $\matr{B} = \left[ \bm b_1 \; \cdots \; \bm b_n \right] \in \mathbb{R}^{m_2 \times n}$ is $\matr{A} \odot \matr{B} = \left[ \bm a_1 \otimes \bm b_1 \; \cdots \; \bm a_n \otimes \bm b_n \right]$ with $\otimes$ denoting the Kronecker product, 
\begin{equation}
	\begin{array}{cc}
	\matr{J}_{K_{j+1}}(\matr{H}_j) \triangleq
	\left[
	\begin{array}{c}
		\matr{J}_{\kappa_{j+1,1}}(\matr{H}_j) \\
		\matr{J}_{\kappa_{j+1,2}}(\matr{H}_j) \\
		\vdots               \\
		\matr{J}_{\kappa_{j+1,N}}(\matr{H}_j)
	\end{array}
	\right]
	\in \mathbb{R}^{N^2 \times s_j}
& \text{and} \quad
	\matr{J}_{\kappa_{j+1,i}}(\matr{H}_j) \triangleq
	\left[
	\begin{array}{c}
		\nabla k_{j+1}(\bm h_i^{(j)},\bm h_1^{(j)})^\top \\
		\nabla k_{j+1}(\bm h_i^{(j)},\bm h_2^{(j)})^\top \\
		\vdots                                   \\
		\nabla k_{j+1}(\bm h_i^{(j)},\bm h_N^{(j)})^\top
	\end{array}
	\right]
	\in \mathbb{R}^{N \times s_j}.
	\end{array}
\end{equation}}

Below, two examples are illustrated on the derivations of $\matr{\mathcal{G}}$ when different kernel functions are chosen.
\begin{example}[Linear kernel] \label{ex:lin}
	In the case of linear $k_j(\bm z,\bm y)=\bm z^\top \bm y$, we obtain $\matr{J}_{\kappa_{j,i}}=\matr{H}_{j-1}^\top $, so we further have $\matr{\mathcal{G}}_{j-1}=\matr{H}_{j}\matr{H}_{j}^\top \matr{H}_{j-1}$, and $\matr{K}_j(\matr{H}_{j-1})=\matr{H}_{j-1} \matr{H}_{j-1}^\top$. {In two-level architectures with $k_2(\bm z,\bm y)=\bm z^\top \bm y$,  $\matr{\mathcal{G}}_1$ has a linear dependency on $\matr{H}_1$ as $\matr{\mathcal{G}}_1=\matr{H}_2\matr{H}_2^\top \matr{H}_1$, where the eigendecomposition for the first level is written in the form of $\matr{M}_1(\matr{H}_2)\matr{H}_1=\matr{H}_1\matr{\Lambda}_1$ with $\matr{M}_1(\matr{H}_2)$ independent of $\matr{H}_1$.}
\end{example}

\begin{example}[RBF kernel] \label{ex:rbf}
	Consider $k_j(\bm z,\bm y)=\exp\left(-\norm{\bm z-\bm y}^2_2/(2\sigma^2)\right)$. The partial derivative  is
	\begin{equation}
		\dfrac{\partial k_j(\bm h_i^{(j-1)}, \bm h_k^{(j-1)})}{\partial \bm h_i^{(j-1)}} = -\frac{1}{\sigma^2}\left(\bm h_i^{(j-1)}-\bm h_k^{(j-1)}\right) k_j(\bm h_i^{(j-1)}, \bm h_k^{(j-1)}),
	\end{equation}
	so $\matr{J}_{\kappa_{j,i}}=-2\gamma \, \mathrm{diag}({K_j}_{: i}) \left[\bm h_i^{(j-1)} \mathbbm{1}^\top  - \matr{H}_{j-1} \right]  $, where $\mathbbm{1}$ is a vector of all ones and ${K_j}_{: i}$ is the $i$-th column of $\matr{K}_j$.
\end{example}

The derivations to the dual formulations show that $\matr{\Lambda}_j$ relates to the first $s_j$  eigenvalues corresponding to the $s_j$ eigenvectors $\matr{H}_j$ in the optimization of DPKCA, indicating that all the pairs $(\matr{\Lambda}_j, \matr{H}_j), \, j=1,\dots,\nlevels$ solving the dual problem constitute a pool of candidate solutions that lead to $J=0$ in the primal objective \eqref{eq:deeprkm:primal:main}. Thus, the regularization hyperparameters $\matr{\Lambda}_j$ in the primal are automatically determined  in the dual by the solutions of \eqref{eq:system:nl:main}. Such obtained $\matr{H}_j$ and $\matr{\Lambda}_j$ with $j=1, \ldots, \nlevels$ are named as deep eigenvectors and deep eigenvalues in DKPCA, respectively. The dual problem of DKPCA in each level is interpreted as an eigenvalue problem, giving the conjugated hidden features (principal components) $\matr{H}_j$ solved by the deep eigenvectors corresponding to  level $j$.  The existing (shallow) KPCA is a special case of DKPCA with $\nlevels=1$, where $\matr{\Lambda}_1$ degenerates to the first $s_1$ eigenvalues corresponding to the $s_1$  eigenvectors (principal components) $\matr{H}_1$  of the kernel matrix $\matr K_1$.

\subsection{Optimization Algorithm} \label{sec:methods:opt}
\begin{algorithm}[t]
    \caption{DKPCA Training using PGD. The stepsize $\alpha$ is selected via backtracking for each variable.} \label{alg:train}
	\begin{algorithmic}[1]
		\Function{DeepKPCA}{$\{\bm x_i\}_{i=1}^N$, $\varepsilon > 0$}

            \State Compute $\matr{K}_1$ from $\{\bm x_i\}_{i=1}^N$
		\State Initialize $\{\matr{H}^1_1, \dots, \matr{H}^1_{\nlevels}, \matr{\Lambda}^1_{1}, \dots, \matr{\Lambda}^1_{\nlevels} \}$
		\State $k \gets 0$
		\Repeat
		\State $k \gets k+1$
            \State Compute $\matr{K}_j^k$ from $\matr{H}^k_{j-1}, \, \forall j = 2, \dots, \nlevels$
            \State Compute $\matr{\mathcal{G}}_j^k$ from $\matr{H}_j^k, \matr{H}_{j+1}^k, \, \forall j = 1, \dots, \nlevels-1$
            \State Compute the residuals in \eqref{eq:obj:nl:main}
		\State $\matr{H}^{k+1}_j \gets \mathbf{\Pi}_{\text{St}(s_j,N)}\left(\matr{H}^k_j - \alpha_k \nabla_{\matr{H}_j} \widetilde{J} \left(\matr{H}^k_1, \dots, \matr{H}^k_{\nlevels},\matr{\Lambda}^k_{1},\dots,\matr{\Lambda}^{k}_{\nlevels}\right)\right)$  \Comment{Update for all levels}
		\State $\matr{\Lambda}^{k+1}_j \gets \matr{\Lambda}^k_{j} - \alpha_k \nabla_{\matr{\Lambda}_j} \widetilde{J} \left(\matr{H}^k_1,\dots,\matr{H}^k_{\nlevels},\matr{\Lambda}^k_{1},\dots,\matr{\Lambda}^k_{\nlevels}\right)$ \Comment{Update for all levels}
		
		\Until{$\nicefrac{\norm{\matr{H}^{k+1}_j - \matr{H}^k_j}_{\text{max}}}{\alpha_k} \leq \varepsilon$ \textbf{and} $\nicefrac{\norm{\matr{\Lambda}^{k+1}_{j} - \matr{\Lambda}^k_{j}}_{\text{max}}}{\alpha_k} \leq \varepsilon$} \Comment{Condition for all levels}
		
		\State
		\Return $\matr{H}_1, \dots, \matr{H}_{\nlevels}, \matr{\Lambda}_1, \dots, \matr{\Lambda}_{\nlevels}$
		
		\EndFunction
		
	\end{algorithmic}
\end{algorithm}

For general positive definite kernels $k_j$, \eqref{eq:system:nl:main} {is interpreted as a set of eigendecompositions with optimization variables $\matr H_j$ coupled with previous and subsequent layers. In the algorithmic aspect, we propose to train} the DKPCA by residual minimization of \eqref{eq:system:nl:main}, {which considers the  orthogonality constraints on intra-level hidden features and results} in the following constrained optimization problem:
{
\begin{equation} \label{eq:obj:nl:main}
\begin{alignedat}{3}
&\stackunder{minimize}{$\matr{H}_j,\matr{\Lambda}_j$} &\qquad& \widetilde{J} \triangleq \frac{1}{2} \norm{\dfrac{1}{\eta_1} \matr{K}_1 \matr{H}_1 + \dfrac{1}{\eta_2} {\matr{\mathcal{G}}_1(\matr{H}_1,\matr{H}_2)} - \matr{H}_1 \matr{\Lambda}_1}^2_{\fro} + &\\
&&&\sum_{j=2}^{\nlevels-1} \norm{ \dfrac{1}{\eta_j} \matr{K}_j(\matr{H}_{j-1}) \matr{H}_j + \dfrac{1}{\eta_{j+1}} {\matr{\mathcal{G}}_j(\matr{H}_j,\matr{H}_{j+1})} - \matr{H}_j \matr{\Lambda}_j}^2_{\fro} + &\\
&&& \norm{\dfrac{1}{\eta_{\nlevels}} \matr{K}_{\nlevels}(\matr{H}_{\nlevels-1}) \; \matr{H}_{\nlevels} - \matr{H}_{\nlevels} \matr{\Lambda}_{\nlevels}}^2_{\fro} &\\
&\text{subject to} &      & \matr{H}_j^\top \matr{H}_j = \matr{I}_{s_j}, \quad \forall j=1,\dots,\nlevels,  &\\
\end{alignedat}
\end{equation}
where $\widetilde{J}$ denotes the optimization objective and the residual error is adopted as the Frobenius norm $\norm{\cdot}_{\fro}$.}
{During the training, the hidden features not only flow forward from the previous level, but also backward from the subsequent level, as $\matr{H}_j$ comes from the eigendecomposition 
depending on $\matr{H}_{j-1}$ and $\matr{H}_{j+1}$ in  a level-wise fashion.}

The constraint set for the hidden features of level $j$ is the Stiefel manifold $\text{St}(s_j,N)=\{\matr{H}_j \in \mathbb{R}^{N\times s_j} \, | \, \matr{H}_j^{\top} \matr{H}_j=I_{s_j}\}$. Optimization of \eqref{eq:obj:nl:main} can be tackled by the Projected Gradient Descent (PGD) algorithm, where the iterates for $\matr{H}_j$ are specified by
$
	\matr{H}^{k+1}_j = \mathbf{\Pi}_{\text{St}(s_j,N)}\left(\matr{H}^k_j - \alpha_k \nabla_{\matr{H}_j} \widetilde{J}\left(\matr{H}_1^k, \matr{\Lambda}^{k}_1, \dots, \matr{H}_{\nlevels}^k, \matr{\Lambda}_{\nlevels}^k \right)\right)
$, in the $(k+1)$-th iteration,
with $\mathbf{\Pi}_{\text{St}(s_j,N)}$ being the Euclidean projection onto the Stiefel manifold, and $\alpha_k$ is the stepsize selected via backtracking. The projection is computed using the compact SVD of $\matr{H}^k_j$. 
This algorithm is detailed in Algorithm \ref{alg:train}.
{Since PGD requires the SVD of $\matr{H}_j$ at each iteration for the projection, it can be computationally expensive for large $N$ and $s_j$. For this setting, the Riemannian Adam algorithm \cite{becigneul2019} can be an alternative for  this  constrained optimization, where  each iteration is computationally less expensive.}

\subsection{Generative DKPCA} \label{sec:dkpca:gen}

\begin{figure}
    \centering
    \def\svgwidth{\textwidth}
    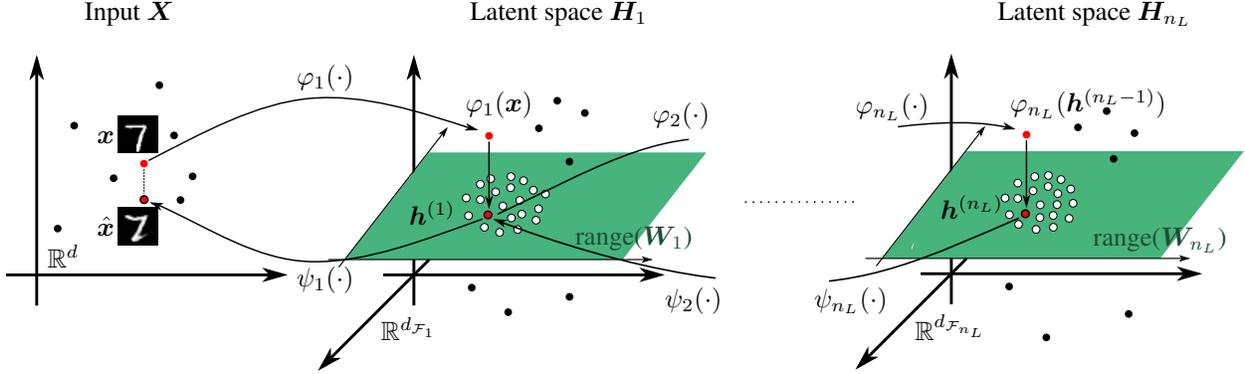
    \caption{Overview of generative DKPCA with $n_L$ levels.
    Multiple latent spaces are considered with multi-level hidden features $\matr{h}^{(j)}, \, \forall j=1,\dots,n_L$.
    The feature maps $\varphi_j$ are indicated with arrows going from left to right. The generative model employs the pre-image maps $\psi_j$, represented by the arrows going from right to left.
    The dashed line in input space represents the reconstruction error.
    The projecting vector in latent spaces indicates the projections in the corresponding $s_j$-dimensional latent subspace.}
    \label{fig:projections}
\end{figure}

In linear PCA, performing reconstruction is straightforward by
a linear basis transformation, while the nonlinear KPCA faces the well-known pre-image challenges in reconstructions \cite{mika1999}.
The proposed DKPCA employs multiple nonlinear feature maps and
consists of multiple latent spaces, posing even greater challenges for the  reconstruction. 
We propose a procedure for 
generative 
DKPCA
from the
sampled 
hidden features $\bm h^{(j)}$ in latent spaces with parametric
feature maps  $\varphi_j$ of each level, which also induces a positive definite kernel matrix \cite{suykens2002,pandey2021}.
{
We also describe how the proposed generative model can facilitate
the exploration of}
the role of the deep eigenvectors of each level.

Given the learned  $\bm h^{(j)}$, we consider a generative objective introducing one term per level to the objective \eqref{eq:deeprkm:primal:main} for a point $\bm x$: $\frac{1}{2} \varphi_1(\bm x)^\top  \varphi_1(\bm x)$ for the first level and $\tfrac{1}{2} \varphi_j\left(\bm h^{(j-1)}\right)^\top  \varphi_j\left(\bm h^{(j-1)}\right)$ for level $j=2,\dots,\nlevels$. By the characterization of the stationary points given in \ref{sec:methods:gen}, a 
new point $\hat{\bm x}$  is generated through the inverse maps of the multiple levels:
\begin{equation}
	\label{eq:genx:main}
	\hat{\bm x} = \varphi_1^{-1} \left( \matr{W}_1 \bm a^{(2)} \right),
\end{equation}
such that $\bm a^{(j)} = \varphi_j^{-1} \left( \matr{W}_j \bm a^{(j+1)} \right),  j=2,\dots,\nlevels$, $\bm a^{(\nlevels+1)} = \bm h^{(\nlevels)}$, and where  $\varphi_j$ is invertible  with the inverse map denoted as $\varphi_j^{-1}$.
Note that \eqref{eq:genx:main} has a similar structure to the decoder of an Autoencoder architecture. This process is visualized in Fig. \ref{fig:projections}.

In practice, it is particularly useful to employ parametric feature maps, as they can learn to well map high-dimensional complex data from the unknown training distribution. For instance, in computer vision tasks one can define a convolutional neural network as the feature map $\varphi_1$. A transposed convolutional network $\psi_1$ is used in the generation formula \eqref{eq:genx:main} to approximate the inverse map $\varphi_1^{-1}$ such that $(\psi_1 \circ \varphi_1)(\bm x) \approx \bm x$.  In such cases when the inverse map $\varphi_1^{-1}$ is unknown explicitly in advance, one can employ a learnable pre-image map to approximate the inverse map, which resembles the decoder part in an  Autoencoder  architecture. Thus, we add the reconstruction error, e.g., $\mathcal{L}_i(\bm x_i, \psi_1(\varphi_1(\bm x_i))) = \norm{\bm x_i - \psi_1(\varphi_1(\bm x_i))}^2$  to the optimization objective  $\widetilde{J}$ in \eqref{eq:obj:nl:main} for the learning of the  inverse feature map $\psi_1$.
The full  objective is thereby cast as 
\begin{equation}  \label{eq:exp:obj:main}
	\widetilde{J} + \gamma \sum_{i=1}^N \mathcal{L}_i\left(\bm x_i, \psi_1(\varphi_1(\bm x_i))\right),
\end{equation}
where {$\psi_1$ is the learnable {pre-image} map that approximates the inverse map  $\varphi_j^{-1}$, $\mathcal{L}_i$ is the reconstruction error of sample $\bm x_i$}, and $\gamma>0$ balances the
reconstruction error and the residuals minimization. Besides $\matr H_j$ and $\matr \Lambda_j$ in  $\widetilde{J}$, the network parameters of $\varphi_1$ and $\psi_1$ also need to be learned. In this optimization problem, an alternating update scheme is adopted: {the Adam optimizer \cite{adam} is used to update the parameters of $\varphi_1$ and $\psi_1$, while keeping the deep eigenvectors and eigenvalues fixed;  the hidden features $\matr H_j$  and the corresponding eigenvalues $\matr \Lambda_j$ are updated using the DKPCA training algorithm described in Section \ref{sec:methods:opt} with $\varphi_1$ and $\psi_1$  fixed.}

In this case, the optimization to the proposed generative model includes both  the latent variables $\bm h^{(j)}_i$ in the dual and  the explicit feature map $\varphi_1$  in the primal.  This combination allows both the couplings of the levels in the latent variables of each level and deep powerful parametric feature maps better suited for more complex tasks. The deep architecture  of DKPCA consists of  feature maps over multiple levels, where depth is given both by multiple KPCA levels and by feature maps possibly consisting of multi-layered neural networks. This generative model resolves the pre-image problem in performing reconstruction and also enables to obtain new data corresponding to any sampling in the multiple latent spaces. For reconstruction, given any input, its hidden features (principal components) in latent spaces are first computed and are then fed to  the inverse feature maps for reconstruction in the original input space. 
For generation, given any sampling in latent spaces, their correspondingly generated samples in the input space can be obtained through  the inverse feature maps using \eqref{eq:genx:main}. This  makes it viable to explore the role of the deep eigenvectors relating to the principal components of each level, i.e., the generation of newly sampled latent variables can be investigated  
 by  changing only one  latent variable (principal component) at a time, performing the traversals over these latent variables.

Besides, DKPCA also pertains the out-of-sample extension, which allows to predict unseen input data without retraining. This property is  of particular interest in large-scale case for such unsupervised settings   \cite{suykens2002},  as a subset of 
$M \ll N$ samples are  used for the efficient training and the rest $N-M$ samples can be predicted through out-of-sample extensions, as detailed in \ref {sec:methods:gen}. In this way, the storage complexities for the kernel matrices and the hidden features matrices of level $j$ decrease from $\mathcal{O}(N^2)$ and $\mathcal{O}(Ns_j)$ to  $\mathcal{O}(M^2)$ and $\mathcal{O}(Ms_j)$, respectively.  
One approach to the subset selection  is to take a random subsample of $M$ data points for the training, which is capable of well balancing both efficiency and accuracy as evaluated in Fig. \ref{fig:dis:large}. One can also use more sophisticated
selection schemes, such as 
the quadratic Renyi entropy \cite{girolami2002} or the leverage score sampling \cite{rudi2018}. The optimal selection strategy is nevertheless data-dependent in practical applications \cite{espinoza2003,fanuel2021}.

\section{Analytical Findings} \label{sec:results:theory}
In this section, first we show that the optimization problem of our method  explicitly formulates a set of nonlinear equations for each level resembling an eigenvalue problem of some matrix $\matr{M}_j$ fusing  the principal components of previous and subsequent levels, i.e., DKPCA introduces not only
\textit{forward} couplings, but also \textit{backward} couplings between the levels.
Further, we illustrate that the additional levels act as a regularization on the first level.
Then, we apply the Eckart-Young theorem to the deep kernel machine for approximation error bounds on the kernel matrix of the given data. 
Finally, we show conditions under which the explained variance of DKPCA
is strictly greater than the one from KPCA.

\subsection{Forward and Backward Couplings between Levels}
The equations in \eqref{eq:system:nl:main} give the level-wise eigendecomposition interpretation of DKPCA, in which the forward and backward couplings between levels are embodied. The first level resembles the eigendecomposition of the regularized kernel matrix of the given data $\matr{M}_1 \triangleq \frac{1}{\eta_1} \matr{K}_1 + \frac{1}{\eta_2} {\matr{\mathcal{G}}_1(\matr{H}_1,\matr{H}_2)} \matr{H}_1^\top$; the last level is the eigendecomposition of the symmetric matrix $ \matr{M}_{\nlevels} \triangleq \frac{1}{\eta_{\nlevels}} \matr{K}_{\nlevels}$; the intermediate levels $j=2,\dots,\nlevels-1$ are related to the eigendecomposition of
\begin{equation}\label{eq:M:j:main}
	\matr{M_j}(\matr{H}_{j-1},\matr{H}_j,\matr{H}_{j+1})\triangleq \dfrac{1}{\eta_j} \matr{K}_j(\matr{H}_{j-1}) + \dfrac{1}{\eta_{j+1}} {\matr{\mathcal{G}}_j(\matr{H}_j,\matr{H}_{j+1})} \matr{H}_j^\top,
\end{equation}
with deep eigenvectors $\matr{H}_j$ and deep eigenvalues $\matr{\Lambda}_j$. Fig. \ref{fig:dual} visualizes this process.

The optimization
of  DKPCA 
discussed in Section \ref{sec:methods:opt}
is interpreted as a set of $\nlevels$ 
eigendecomposition problems, 
each of which ($\matr H_{j}$) depends on the hidden features of both previous ($\matr H_{j-1}$) and subsequent ($\matr H_{j+1}$) levels. In this way, information not only flows forward but also backward in the learning process, as $\matr{M}_j$ has dependency on both $\matr{H}_{j-1}$ and $\matr{H}_{j+1}$. This is an important property, as previous theoretical works in deep learning such as \cite{bfc} stressed that forward propagation alone in a level-wise fashion is not enough to learn efficient deep architectures, as the levels also need to be coupled in backward directions so that 
more abstract representation of subsequent levels can be utilized to improve the learning of the current level. With the forward and backward couplings between levels,  eigenvalue problems in  \eqref{eq:system:nl:main} cannot be independently solved in series, which motivates the DKPCA training algorithm by residual minimization of the set of  nonlinear equations \eqref{eq:system:nl:main}
described in Section \ref{sec:methods:opt}.

\subsection{Deep Approximation Analysis}\label{sec:results:analysis}

\begin{figure}[t]
	\centering
	\begin{subfigure}[b]{0.33\textwidth}
		\centering
        \captionsetup{oneside,margin={0cm,-1cm}}
		\includegraphics[width=\textwidth]{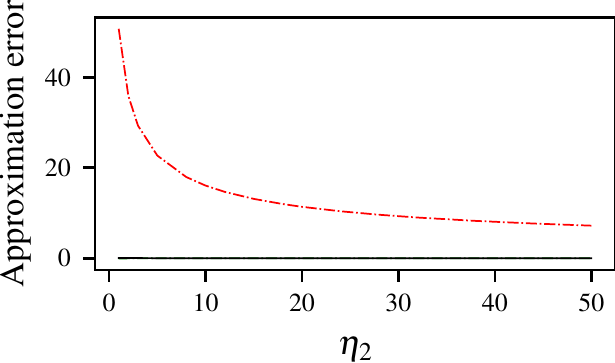}
		\caption{}
		\label{fig:lemmabounds:eta2pos}
	\end{subfigure}
	\hspace{1.5cm}
	\begin{subfigure}[b]{0.33\textwidth}
		\centering
        \captionsetup{oneside,margin={0cm,-1cm}}
		\includegraphics[width=\textwidth]{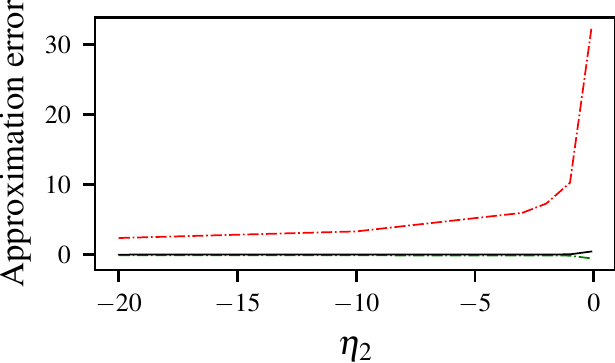}
		\caption{}
		\label{fig:lemmabounds:eta2neg}
	\end{subfigure}
	\\
	\begin{subfigure}[b]{0.33\textwidth}
		\centering
        \captionsetup{oneside,margin={0cm,-1cm}}
		\includegraphics[width=\textwidth]{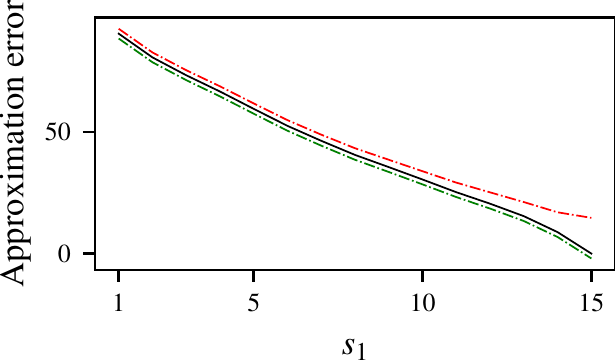}
		\caption{}
		\label{fig:lemmabounds:s1}
	\end{subfigure}
	\hspace{1.5cm}
	\begin{subfigure}[b]{0.33\textwidth}
		\centering
        \captionsetup{oneside,margin={0cm,-1cm}}
		\includegraphics[width=\textwidth]{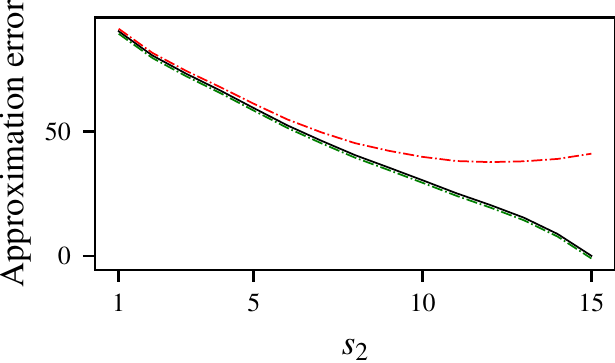}
		\caption{}
		\label{fig:lemmabounds:s2}
	\end{subfigure}
	\caption{\textbf{Deep approximation bounds.} Illustration of Lemma \ref{lemma:lb:ub}. Lower (green dashed line) and upper bound (red dashed line) for the deep approximation error (black solid line) of $\matr{K}_1$ on Synth 3, for varying $\eta_2$ when (\subref{fig:lemmabounds:eta2pos}) $\eta_2>0$ and when (\subref{fig:lemmabounds:eta2neg}) $\eta_2 < 0$, (\subref{fig:lemmabounds:s1}) $s_1$, and (\subref{fig:lemmabounds:s2}) $s_2$. In (\subref{fig:lemmabounds:eta2pos})(\subref{fig:lemmabounds:eta2neg}), the full decomposition case is considered.}
	\label{fig:lemmabounds}
\end{figure}

For theoretical analysis, we consider the two-level DKPCA with $k_2(\matr{z},\matr{y})=\matr{z}^\top \matr{y}$, as the optimization can be
simplified. In this case, $\matr{M}_j$ does not depend on $\matr{H}_j$ such that $\matr M_1(\matr H_2) \matr H_1=\matr H_1 \matr \Lambda_1$ and $\matr M_2(\matr H_1)\matr H_2=\matr H_2 \matr \Lambda_2$, where $\matr{H}_1$ and $\matr{H}_2$ are implemented as
 the eigenvectors in  Level 1 and Level 2, respectively:

\begin{equation}
	\label{eq:system:main}
	\left\{
	\begin{array}{ll}
		\vspace{0.1cm}
		\text{Level 1: } \left( \dfrac{1}{\eta_1} \matr{K}_1 + \dfrac{1}{\eta_2} \matr{H}_2 \matr{H}_2^\top  \right) \matr{H}_1 & = \matr{H}_1 \matr{\Lambda}_1, \\
		
		\text{Level 2: } \left( \dfrac{1}{\eta_2} \matr{H}_1\matr{H}_1^\top  \right) \matr{H}_2                          & = \matr{H}_2 \matr{\Lambda}_2,
	\end{array}
	\right.
\end{equation}
where the first level performs KPCA of $\tfrac{1}{\eta_1} \matr{K}_1 + \tfrac{1}{\eta_2} \matr{H}_2 \matr{H}_2^\top$ and the second level performs KPCA of $\tfrac{1}{\eta_2} \matr{H}_1 \matr{H}_1^\top$. Here,
the second level can be regarded as playing a regularization role: the second level leads to a regularized $\matr{K}_1$ with the regularization constant $\tfrac{\eta_1}{\eta_2}$. Note that $\matr{H}_2$ is unknown a priori, so one has to solve the sets of nonlinear equations \eqref{eq:system:main} for both levels rather than first solving the eigenvalue problem for level 1 and then for level 2, reflecting the forward and backward dependency.

We analyze approximation error bounds for the conceived two-level architectures through the Eckart-Young theorem \cite{eckart1936}, as both of the matrices to be factorized are symmetric, providing additional insights into the DKPCA.

\begin{lemma}[Error bounds] \label{lemma:lb:ub}
	Applying the Eckart-Young theorem to both levels in \eqref{eq:system:main} with orthonormality constraints, the following bound for the deep approximation of $\matr{K}_1$ is obtained
    \begin{equation}  \label{eq:lb:ub}
    \sqrt{\sum_{i=s_1+1}^{r_1} {\lambda_{i}^{(1)}}^2} - \frac{\sqrt{s_2}}{|\eta_2|} \leq
    \norm{\matr{K}_1-\matr{H}_1\matr{\Lambda}_1\matr{H}_1^\top }_F  \leq
    \begin{cases} 
    { \sqrt{ \sum\limits_{i=s_1+1}^{r_1} {\lambda_i^{(1)}}^2 - \left( \frac{s_2}{\eta_2} + 2\sum_{i=1}^{s_2}\widetilde{\lambda}_i \right) \frac{1}{\eta_2}}} & \text{$\eta_2 < 0$},
    \\
    \sqrt{ \sum\limits_{i=s_1+1}^{r_1} {\lambda_i^{(1)}}^2 - \left( \frac{1}{\eta_2} - 2s_1\sum_{i=1}^{s_1}\lambda_i^{(1)} \right)\frac{s_2}{\eta_2}} & \text{$\eta_2 > 0$},
       \end{cases}
    \end{equation}
with $s_1 \leq r_1$, where $r_1=\text{rank}(\matr{K}_1+\tfrac{1}{\eta_2}\matr{H}_2\matr{H}_2^\top )$ and $\widetilde{\lambda}_i$ is the $i$-th largest eigenvalue of $\matr{K}_1$.
\end{lemma}

 Lemma \ref{lemma:lb:ub} gives the error of approximating the data kernel matrix $\matr{K}_1$ with the low-rank matrix of hidden features  $\matr{H}_1$ of the first level as a lower bound depending on the remaining eigenvalues of $\matr{K}_1$ regularized with the matrix of hidden features $\matr{H}_2$ of the second level. The smaller $\eta_2$, the greater the  effect of the second level. On the other hand,  a very large $\eta_2$ indicates high regularization on the second level, reducing its effect, in which the deep architecture behaves resembling a shallow low-rank approximation.
If the number of columns $s_1$ of the approximating matrix is greater than the rank $r_1$ of the matrix to be approximated, one can choose $s_1=r_1$ achieving an error-free approximation. See Fig. \ref{fig:lemmabounds} for numerical evaluation and \ref{sec:methods:theory:bounds} for the proof.

In the next Lemma, we study the cumulative explained variance given by the principal components of the considered two-level DKPCA with comparisons to shallow KPCA, analytically showing the higher explained variance of DKPCA.

\begin{lemma}[Explained variance of deep KPCA]
\label{lemma:explainedvar:lin}
	In the full decomposition case ($s_1=s_2=N$), when $\eta_2 < -\dfrac{1}{\widetilde{\lambda}_{N}}$, the explained variance of the top $n$ principal components of DKPCA in \eqref{eq:system:main} is strictly greater than the variance explained by the top $n$ principal components of shallow kernel PCA, i.e.,
	\begin{equation} \label{eq:var}
		    \frac{\sum_{j=1}^n {\lambda}_j}{\sum_{i=1}^N {\lambda}_i} > \frac{\mya}{\myb},
		\end{equation}
	where $\widetilde{\lambda}_i>0$ is the $i$-th largest eigenvalue of the kernel matrix $\matr{K}_1$, which is taken positive-definite, and $\lambda_i$ is the $i$-th largest eigenvalue of $\matr{K}_1+\frac{1}{\eta_2}\matr{H}_2\matr{H}_2^\top $, for all $ 1\leq n<N$. 

\end{lemma}

The above Lemma gives conditions on $\eta_2$ under which the considered two-level DKPCA is advantageous compared to shallow KPCA in terms of the explained variance of the first $n$ principal components. When choosing $\eta_2 < -\frac{1}{\widetilde{\lambda}_N}$, where $\widetilde{\lambda}_N$ is the smallest eigenvalue of the data kernel matrix, the cumulative variance explained by the first $n$ components of the first DKPCA level is strictly greater than the variance explained by the first $n$ components of shallow KPCA. In other words, DKPCA can capture more information in fewer components. See the next Section for associated numerical experiments and \ref{sec:methods:theory:explainedvar} for the proof.

\section{Numerical Experiments}\label{sec:numerical:test}
We present a series of  experiments to assess and explore  DKPCA, showing the efficacy and advantages of the proposed deep method from different aspects in the following subsections.  DKPCA is implemented in Python using the PyTorch library. The  code is available  at \url{https://github.com/taralloc/deepkpca}, where all datasets used in this study and the setup details are publicly available and  described in the repository.

\paragraph{\textbf{Datasets}} Both synthetic and real-world data are used  to assess  the proposed method with empirical evidence. Three synthetic datasets are presented: a 2D square  dataset (Synth 1), a complex 2D  dataset consisting of one square, two spirals and one ring (Synth 2), and a 140-dimensional multivariate Normal dataset (Synth 3), where samples are drawn randomly from mixed Gaussian distributions. %
For real-world data, we consider  MNIST \cite{mnist}, 3DShapes \cite{factorvae}, Cars3D \cite{cars3d}, and SmallNORB \cite{lecun2004}. In particular, we evaluate disentanglement on the   3DShapes, Cars3D, and SmallNORB, which are popular benchmarks for evaluating variation factors.

\paragraph{\textbf{Evaluation metrics and compared methods}} Different related unsupervised learning methods are adopted to comprehensively evaluate  DKPCA. A comparison to the shallow KPCA is presented with the learned principal components on multiple aspects.   We also consider the state-of-the-art methods $\beta$-VAE \cite{betavae}, FactorVAE \cite{factorvae}, and $\beta$-TCVAE \cite{mig} for general disentangled feature learning.  We keep the same encoder $\varphi_1$ and decoder $\psi_1$ architecture for all  methods. For quantitative evaluations, we employ the IRS metric \cite{irs}, where a higher value indicates better robustness to changes in variation factors. The shared hyperparameters among all methods are fixed to be the same. For the model-specific hyperparameters, we used the suggested values in their papers. 
It is worth mentioning that
the compared methods are sensitive to hyperparameter selections, as shown in \cite{locatello}. 
Our method does not suffer from such  issue as $\matr{\Lambda}_j$ is automatically determined by the solution of the deep KPCA problem and $\eta_j$ is a scaling factor fixed to 1.  More details of the setups  are given in  \ref{sec:app:expsetups}.

\subsection{DKPCA Provides Interpretable Deep Principal Components} \label{sec:results:dis}
This part examines
the roles of each individual deep principal component and of the components  in each level.  Contrary to shallow KPCA owning one set of eigenvectors/eigenvalues, DKPCA have multiple sets of eigenvectors/eigenvalues  for  each level. Thus, the features can be represented in a more hierarchical way that benefits the interpretation explorations. 
In fact,
via the proposed deep generative procedure \eqref{eq:genx:main}, 
sampled hidden features and their pre-image mappings to the input space can be computed.
By traversing the latent space in some specific dimensions,
i.e., varying a single deep principal component  while keeping the others fixed and generating the corresponding sample in input space,
what each component learns can be observed. In DKPCA, with the extracted deep eigenvectors $\matr H_j$, the model can well disentangle the factors of variation in the data.
This is verified quantitatively and qualitatively, comparing the traversals on the learned principal components with the state-of-the-art FactorVAE. 

Notably, 
we show that 
DKPCA 
effectively facilitates hierarchical data exploration, as the role of each principal component in each level can be investigated through the generation of new data. Specifically, we consider images of 3D objects with different generative factors, i.e., colors, sizes, etc. For individual components, our method can find new principal components such that, when sampling along one of them, only
 one generative factor changes, e.g., only the object scale changes,
 while its color and other factors
remain fixed. For the components in each level, our deep method creates a learning hierarchy: 
prevailing features are typically learned 
in the shallower levels, e.g., colors, while the deeper levels capture more subtle features, e.g., the
specific object shape. 

\begin{figure}[t!]
\centering
\begin{subfigure}[b]{\textwidth}
\centering
         \begin{subfigure}[b]{0.05\textwidth}
     \begin{tikzpicture}
      \draw [decorate, decoration = {calligraphic brace}, thick] (0,0) -- (0,2.5)  node [midway,left] {\small $\bm h^{(2)}$}; %
      \draw [decorate, decoration = {calligraphic brace}, thick] (0,2.6) -- (0,3.7) node [midway,left] {\small $\bm h^{(1)}$}; %
     \end{tikzpicture}
     \end{subfigure}
     \begin{subfigure}[b]{0.45\textwidth}
         \centering
         \stackinset{c}{}{t}{-1in}{\stackinset{c}{}{t}{-.2in}{Deep KPCA}{%
         \includegraphics[width=0.65\textwidth]{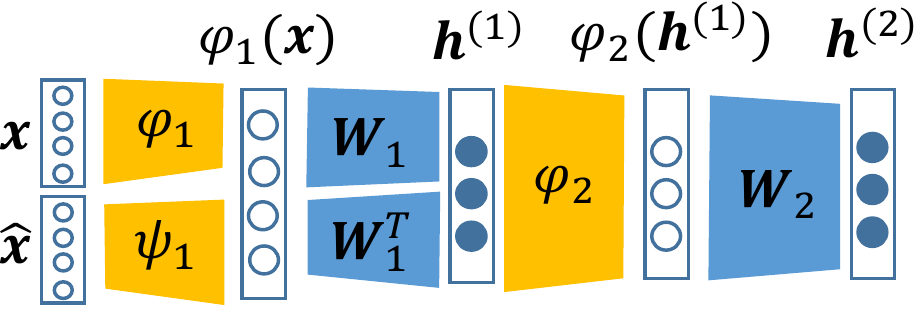}
         }}{%
         \includegraphics[width=\textwidth]{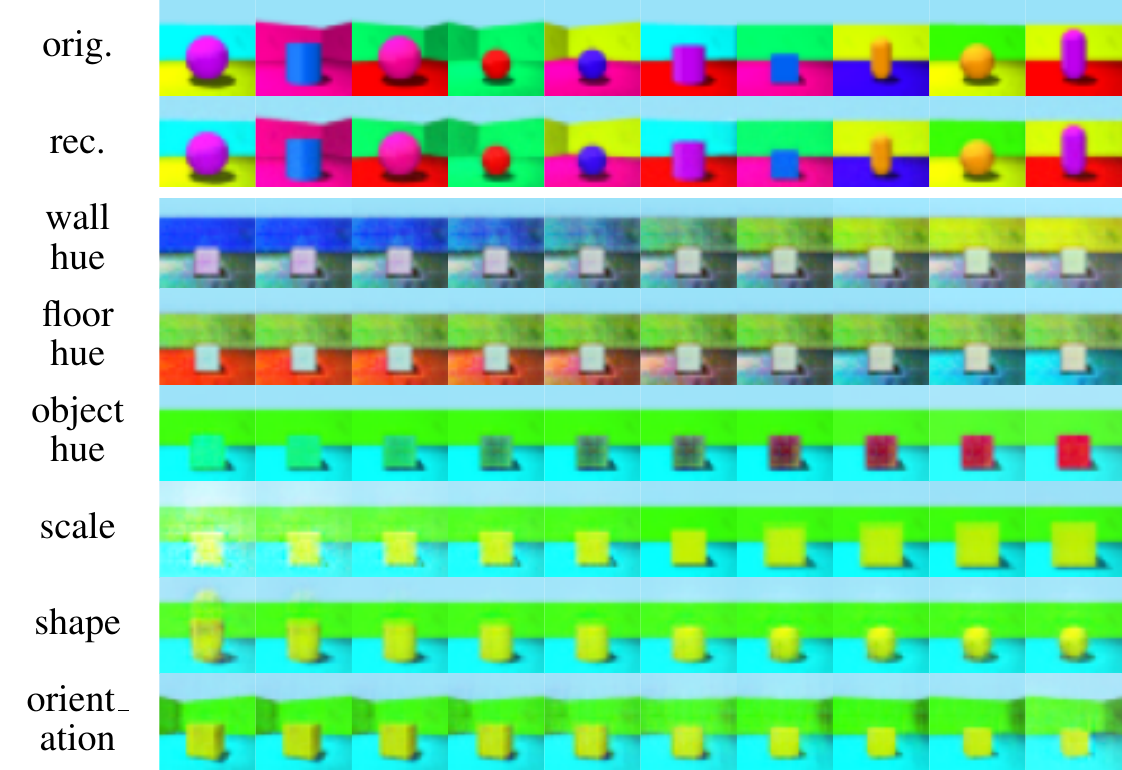}
         }
     \end{subfigure}
     \hfill
      \begin{subfigure}[b]{0.025\textwidth}
     \begin{tikzpicture}
     \draw [decorate, decoration = {calligraphic brace}, thick] (0,0) -- (0,3.7)  node [midway,left] {\small $\bm z$}; %
     \end{tikzpicture}
     \end{subfigure}
      \begin{subfigure}[b]{0.45\textwidth}
         \centering
         \stackinset{c}{}{t}{-1in}{\stackinset{c}{}{t}{-.2in}{FactorVAE}{%
         \includegraphics[width=0.35\textwidth]{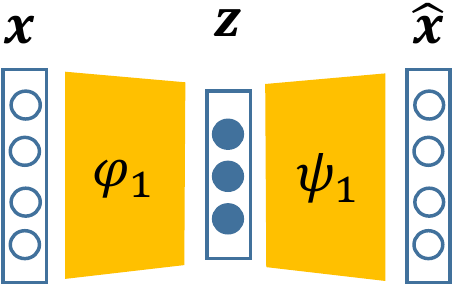}
         }}{%
         \includegraphics[width=\textwidth]{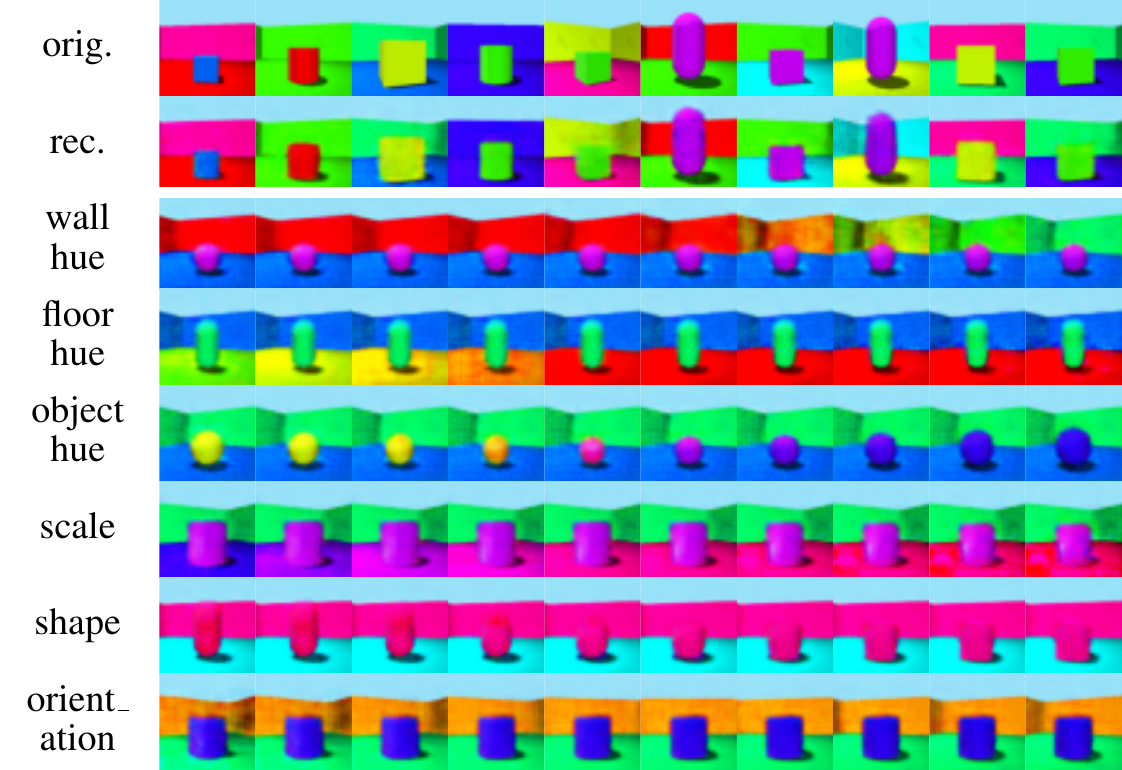}
         }
     \end{subfigure}%
    \caption{}
    \label{fig:main:traversal}
\end{subfigure}%
\\ \vspace{0.3cm}
\begin{subfigure}[b]{0.38\textwidth}
     \centering
       \captionsetup{oneside,margin={0cm,-0.8cm}}
    \includegraphics[width=\textwidth]{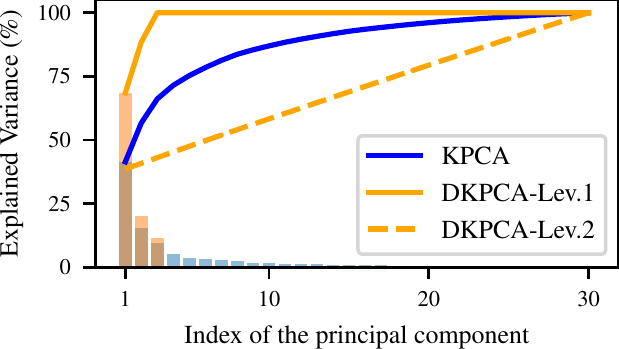}
    \caption{}
    \label{fig:main:explained}
\end{subfigure}%
\hfill
\begin{subfigure}[b]{0.2\textwidth}
\centering
  \captionsetup{oneside,margin={0cm,-0.8cm}}
  \includegraphics[width=\textwidth]{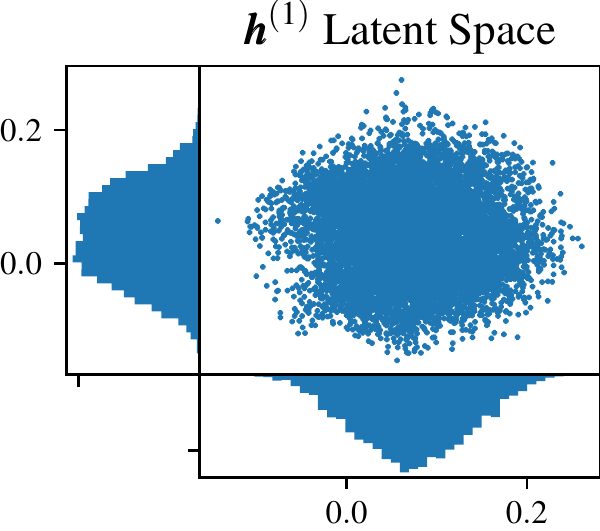}
  \vspace{0.15cm}
  \caption{}
\end{subfigure}%
\hfill
\begin{subfigure}[b]{0.2\textwidth}
\centering
  \captionsetup{oneside,margin={0cm,-0.8cm}}
  \includegraphics[width=\textwidth]{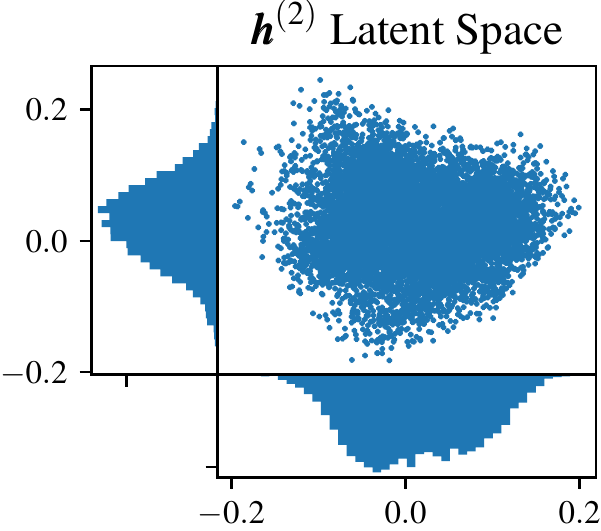}
  \vspace{0.15cm}
  \caption{}
\end{subfigure}%
\hfill
\begin{subfigure}[b]{0.2\textwidth}
\centering
  \captionsetup{oneside,margin={0cm,-0.8cm}}
  \includegraphics[width=\textwidth]{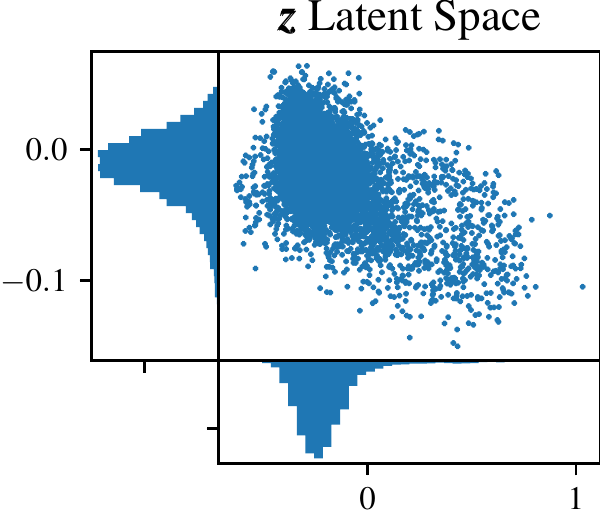}
  \vspace{0.15cm}
  \caption{}
\end{subfigure}
\caption{\textbf{Results on the 3DShapes dataset.} \textbf{(a) Role of the deep principal components.} The ground-truth on the 1st row, reconstructions on the 2nd row, and  traversals on other rows in the latent spaces induced by DKPCA and FactorVAE. The factors extracted by DKPCA are better disentangled than FactorVAE. Unlike FactorVAE, DKPCA shows a hierarchy of details, where the second level learns more complex factors of variation than the first level. \textbf{(b) Explained variance} (\%)  of both DKPCA and shallow KPCA using the same kernel. DKPCA captures considerably greater explained variance (informative features) in the first principal components than KPCA, where the lines denote the cumulative explained variance and the bars denote the variance explained by each component. \textbf{(c)(d)(e)  Scatter plots of the latent variable distribution}, where DKPCA learns one latent space for each level. The FactorVAE distribution shows partial irregularity, while the distributions learned by DKPCA follow a more compact Gaussian profile, centered around the origin in the second level.}
\label{fig:main}
\end{figure}

Fig. \ref{fig:main} summarizes the main results for 3DShapes. Detailed analysis is given in the following for each principal component in all levels and for each level separately.

\begin{figure}[ht!]
\centering
\begin{subfigure}[b]{\textwidth}
         \begin{subfigure}[b]{0.05\textwidth}
     \begin{tikzpicture}
     \draw [decorate, decoration = {calligraphic brace}, thick] (0,0) -- (0,1.2)  node [midway,left] {\small $\bm h^{(2)}$}; %
      \draw [decorate, decoration = {calligraphic brace}, thick] (0,1.3) -- (0,1.8) node [midway,left] {\small $\bm h^{(1)}$}; %
     \end{tikzpicture}
     \end{subfigure}
     \begin{subfigure}[b]{0.45\textwidth}
         \centering
         \stackinset{c}{}{t}{-.2in}{\stackinset{c}{}{t}{-.2in}{Deep KPCA}{%
         }}{%
         \includegraphics[width=\textwidth]{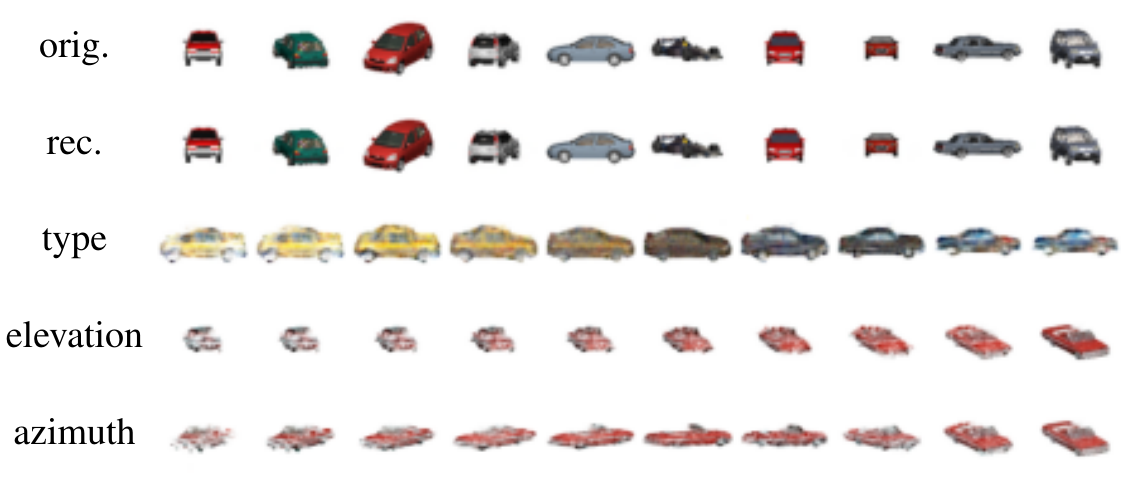}
         }
     \end{subfigure}
     \hfill
      \begin{subfigure}[b]{0.025\textwidth}
     \begin{tikzpicture}
     \draw [decorate, decoration = {calligraphic brace}, thick] (0,0) -- (0,1.8)  node [midway,left] {\small $\bm z$}; %
     \end{tikzpicture}
     \end{subfigure}
      \begin{subfigure}[b]{0.45\textwidth}
         \centering
         \stackinset{c}{}{t}{-.2in}{\stackinset{c}{}{t}{-.2in}{FactorVAE}{%
         }}{%
         \includegraphics[width=\textwidth]{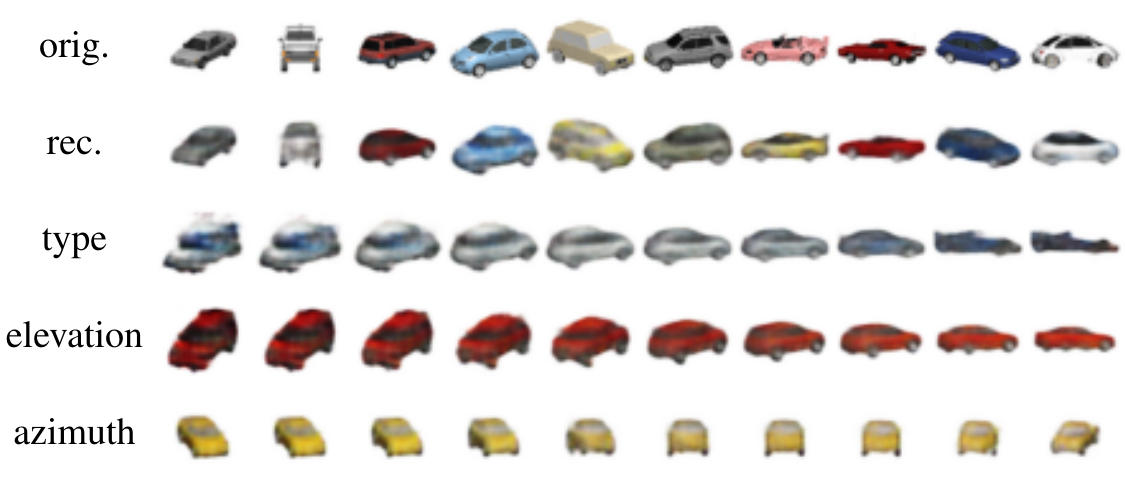}
         }
     \end{subfigure}%
\caption{Cars3D}
\label{fig:dis:cars3d}
\end{subfigure}
\\ \vspace{0.3cm}
\begin{subfigure}[b]{\textwidth}
     \begin{subfigure}[b]{0.05\textwidth}
     \begin{tikzpicture}
     \draw [decorate, decoration = {calligraphic brace}, thick] (0,0) -- (0,1.8)  node [midway,left] {\small $\bm h^{(2)}$}; %
      \draw [decorate, decoration = {calligraphic brace}, thick] (0,2) -- (0,2.5) node [midway,left] {\small $\bm h^{(1)}$}; %
     \end{tikzpicture}
     \end{subfigure}
    \begin{subfigure}[b]{0.45\textwidth}
         \centering
         \includegraphics[width=\textwidth]{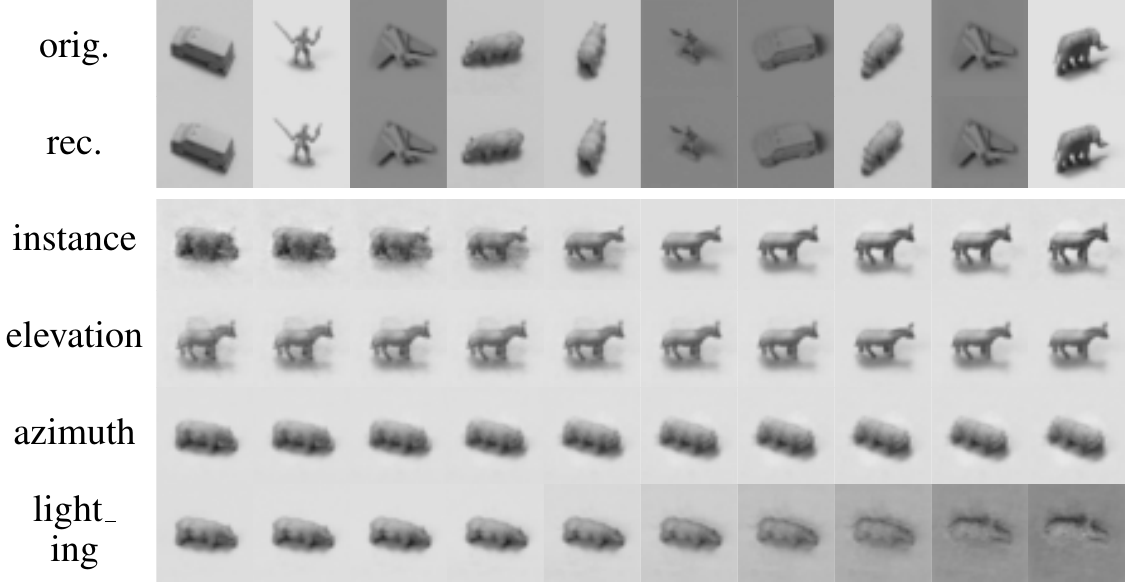}
     \end{subfigure}
     \hfill
     \begin{subfigure}[b]{0.025\textwidth}
     \begin{tikzpicture}
     \draw [decorate, decoration = {calligraphic brace}, thick] (0,0) -- (0,2.5)  node [midway,left] {\small $\bm z$}; %
     \end{tikzpicture}
     \end{subfigure}
      \begin{subfigure}[b]{0.45\textwidth}
         \centering
         \includegraphics[width=\textwidth]{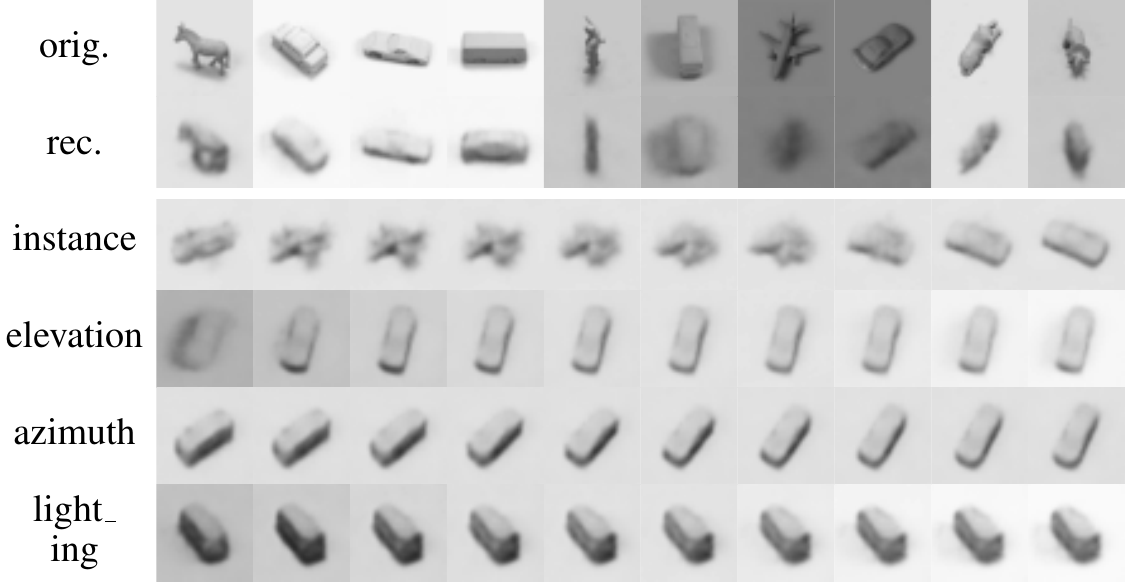}
     \end{subfigure}
 \caption{SmallNORB}
 \label{fig:dis:norb}
\end{subfigure}     
	\caption{\textbf{Role of the deep principal components}. First row: ground-truth. Second row: reconstructions. Other rows: traversals in the latent spaces induced by DKPCA. The DKPCA shows a hierarchy of details, where the second level learns more complex factors of variation than the first level.
	}
	\label{fig:dis}
\end{figure}

\paragraph{\textbf{Individual principal components}} 
In Fig. \ref{fig:main:traversal} and \ref{fig:dis}, we show the traversals in the latent spaces of a DKPCA with explicit feature maps. 
Aside  of high visual reconstruction quality  in the second row,  other rows 
show the generated images while traversing along the individual principal component of the first level ($\matr{h}^{(1)}$) or of the second level ($\matr{h}^{(2)}$) of the proposed DKPCA that explains the corresponding generative factor. In FactorVAE, a single latent space is obtained, {and} the  images are generated by traversing
along each dimension in the latent space of FactorVAE.
For instance, in 3DShapes (Fig. \ref{fig:main:traversal}), the component in Row 3 captures the factor of wall hue, as both the floor and object hue remain almost constant. 
In 3DShapes,  DKPCA better disentangles the scale of the object, 
which only slightly varies in FactorVAE.
In Cars3D (Fig. \ref{fig:dis:cars3d}), the three factors of elevation,  car type, and azimuth by 
DKPCA are well captured and disentangled, while FactorVAE gives entanglement in differentiating the learning of azimuth with  the two other components of elevation and  car type. 
A similar analysis is conducted for the other rows, showing that the deep components well capture the factors of variation  of the data. Besides, thanks  to the eigenvalues $\bm \Lambda_j$ obtained in the optimization, DKPCA can identify an ordering of the components, providing a way to reflect their relative importance. 
This cannot be done with the considered VAE-based methods \cite{betavae,factorvae,mig}.

\paragraph{\textbf{Principal components in each level}} Besides individual components,
we further explore the level-wise 
interpretation of the learned deep principal components in DKPCA. 
In Fig. \ref{fig:main:traversal},
the two components of the first level capture the background, which corresponds to the factors of the highest variation, i.e., the wall and floor hue, as they involve the most pixels in the images. 
The two components of the second level capture subtle characteristics of the object, e.g., scale and orientation, as
the deeper components 
capture generative factors for more {detailed} information with less variation among samples. In other words, DKPCA learns a hierarchy of abstraction in its deep components, from less abstract, i.e., background, to more abstract, i.e., object.
Similar conclusions hold for Cars3D (Fig. \ref{fig:dis:cars3d}): the first level  learns the car type, which is the factor of highest variation, while the second level learns more sophisticated factors capturing the elevation and azimuth of the car.

\begin{figure}[t!]
	\centering
	\begin{subfigure}[b]{0.32\textwidth}
		\centering
		\includegraphics[width=\textwidth]{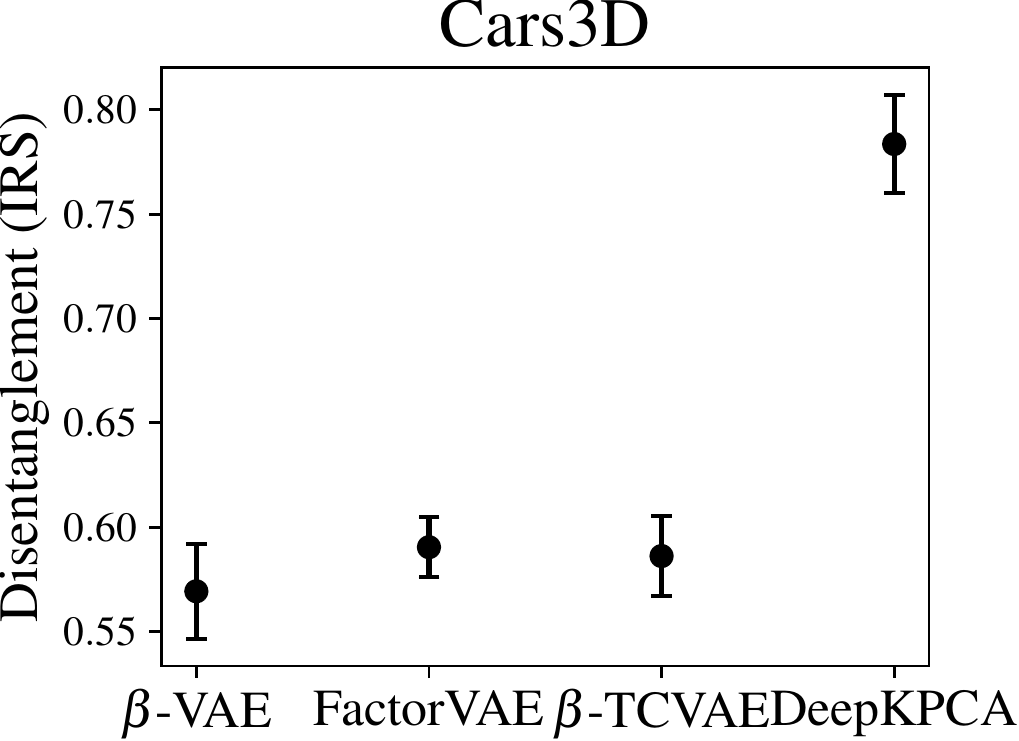}
		\label{fig:dis:small:cars3d}
	\end{subfigure}
	\hfill
	\begin{subfigure}[b]{0.32\textwidth}
		\centering
		\includegraphics[width=\textwidth]{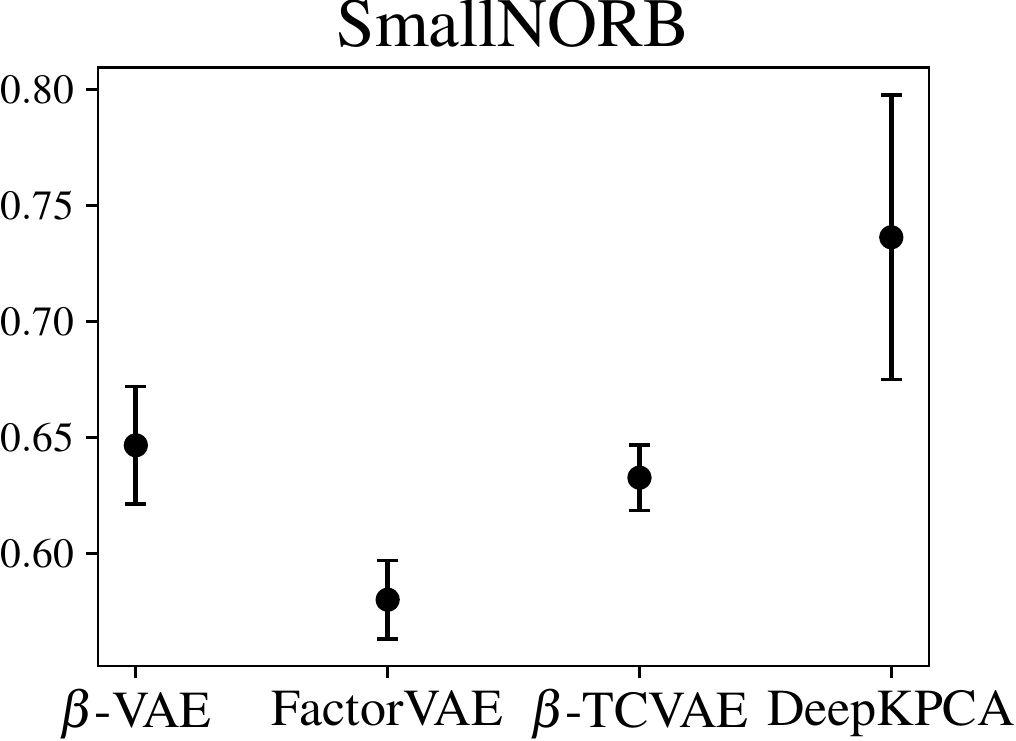}
		\label{fig:dis:small:norb}
	\end{subfigure}
	\hfill
	\begin{subfigure}[b]{0.32\textwidth}
		\centering
		\includegraphics[width=\textwidth]{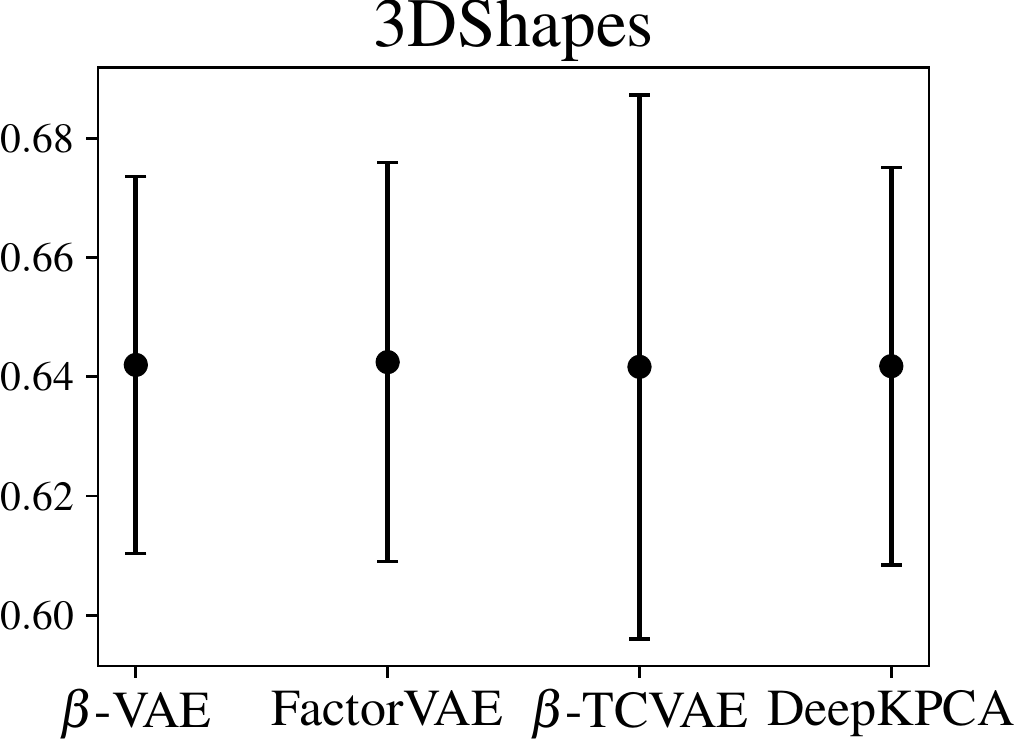}
		\label{fig:dis:small:3dshapes}
	\end{subfigure}
	\caption{\textbf{Disentanglement with small data ($\uparrow$).} 
	Distribution (mean and standard deviation) of disentanglement scores (IRS) for different methods with $N=200$ samples. Higher is better ($\uparrow$).}
	\label{fig:dis:small}
\end{figure}

\paragraph{\textbf{Disentanglement learning}} A quantitative evaluation of disentangled feature learning is performed by comparing with the state-of-the-art methods $\beta$-VAE \cite{betavae}, FactorVAE \cite{factorvae}, and $\beta$-TCVAE \cite{mig} on the commonly used IRS metric \cite{irs}. The studied DKPCA architecture has $\nlevels=2$, $s_1=s_2$ set to the 
number of generative factors, and the latent representation of a data point $\bm x_i$ is given by the concatenation of $\bm h^{(1)}_1$ and $\bm h^{(2)}_2$. The dimension of the latent space of the compared methods is set to $s_1+s_2$. Fig. \ref{fig:dis:small} gives the performance evaluation with models trained on a subset of $N=200$ samples.
The proposed DKPCA shows overall favorable performance  for disentanglement learning on the
tested datasets, notably outperforming the state-of-the-art VAE-based methods in Cars3D. Those advantageous results of DKPCA  achieved under this setting reflects 
better sample efficiency in this set of experiments: from only hundreds of data points, the DKPCA can learn more disentangled representations than the compared data-hungry deep learning methods.
In real-life scenarios, this property can be of particular interest, as the training examples might be available in limited quantity or  expensive to collect, so models  better capturing the true generative factors from 
a limited number of data are desirable.

\begin{figure}[t!]
	\centering
	\begin{subfigure}[b]{0.32\textwidth}
		\centering
		\includegraphics[width=\textwidth]{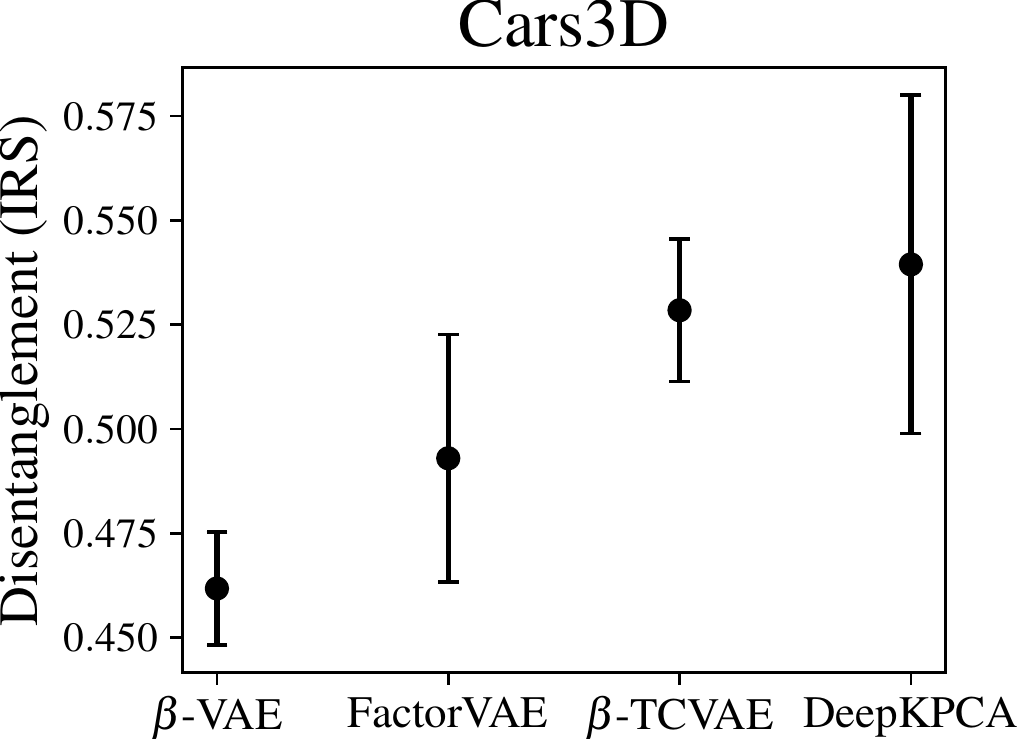}
		\label{fig:dis:large:cars3d}
	\end{subfigure}
	\hfill
	\begin{subfigure}[b]{0.32\textwidth}
		\centering
		\includegraphics[width=\textwidth]{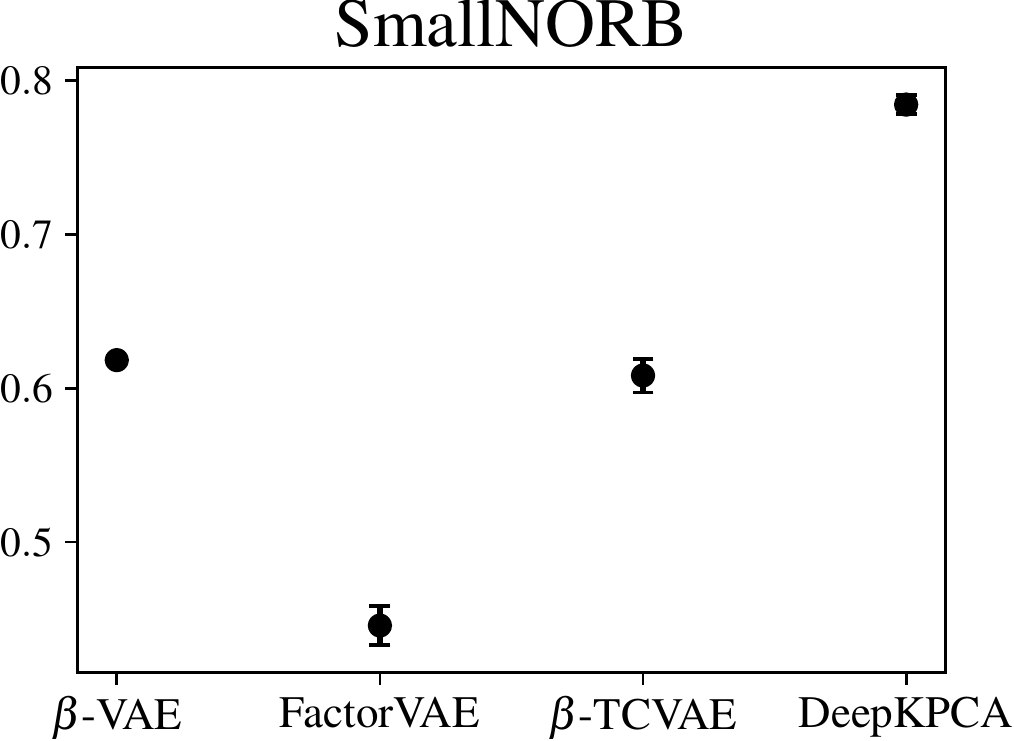}
		\label{fig:dis:large:norb}
	\end{subfigure}
	\hfill
	\begin{subfigure}[b]{0.32\textwidth}
		\centering
		\includegraphics[width=\textwidth]{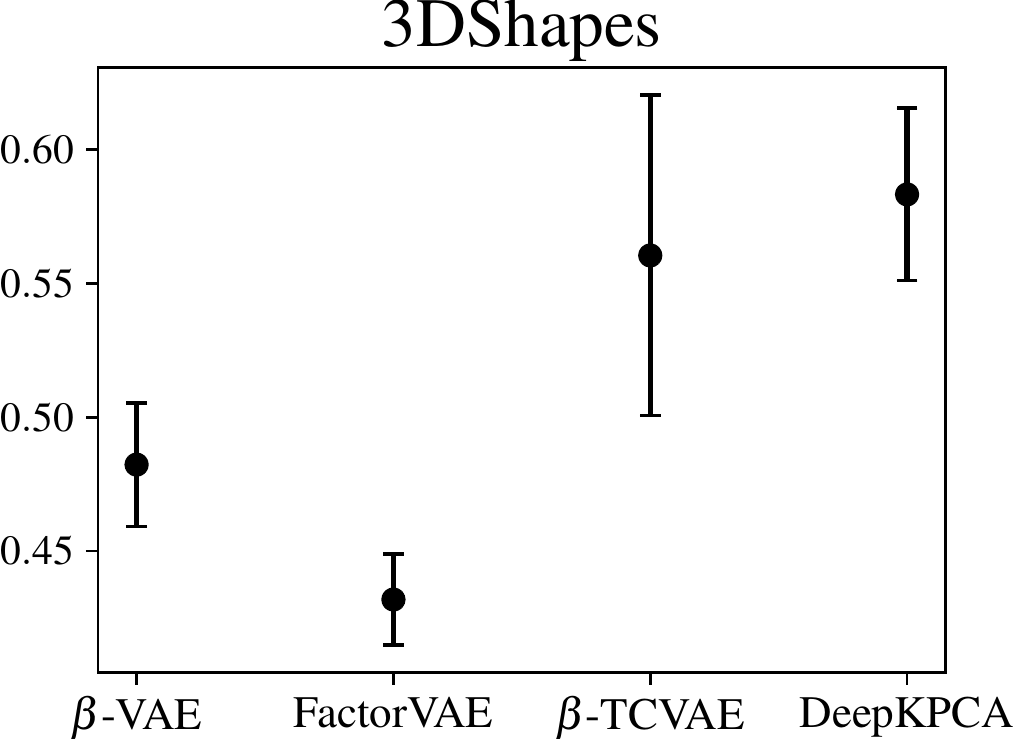}
		\label{fig:dis:large:3dshapes}
	\end{subfigure}
	\caption{\textbf{Disentanglement with large-scale extensions ($\uparrow$).} Disentanglement score (IRS) on the full datasets. A subset of $M=200$ is selected for training our method and the remaining data is inferred with out-of-sample extensions, which is particularly beneficial under limited computational resources.}
	\label{fig:dis:large}
\end{figure}

DKPCA can be implemented with out-of-sample extensions for large-scale cases by selecting a subset $M \ll N$ for training and then 
obtaining
the latent representations of the remaining data.
To evaluate the performance of DKPCA on the full dataset, in Fig. \ref{fig:dis:large} we evaluate  the entire corresponding datasets through out-of-sample extensions using $M=200$ samples for the training. The results shows a higher mean IRS is attained over
all compared methods which are
trained on the full dataset. This comparison further verifies the  disentanglement of the hidden features learned by our method, as well as its sample efficiency: only hundreds of samples are needed by DKPA to effectively learn disentangled representations and outperform the deep learning methods trained on thousands of data points.

\subsection{DKPCA Learns More Informative Features} \label{sec:results:eigs}

In this section, we further investigate the features learned by DKPCA. 
DKPCA gives higher explained variance  than shallow KPCA, indicating that more information is captured in fewer components.
We therefore show the superiority of DKPCA as a feature extractor for downstream supervised tasks %
for multiple data types. 
We also investigate the problem of selecting the number of principal components in each level and the number of levels, providing a selection strategy in an unsupervised setting, in contrast with typical trial and error tuning in deep learning.

\begin{figure}[t!]
	\centering
	\begin{subfigure}[b]{0.32\textwidth}
		\centering
		\includegraphics[width=\textwidth]{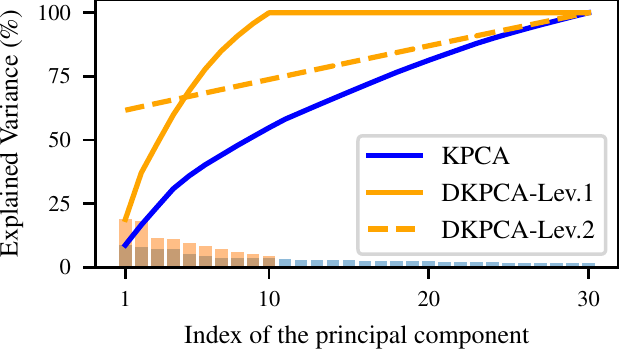}
		\caption{Synth 2.}
		\label{fig:eigs:synth2}
	\end{subfigure}
	\hfill
	\begin{subfigure}[b]{0.32\textwidth}
		\centering
		\includegraphics[width=\textwidth]{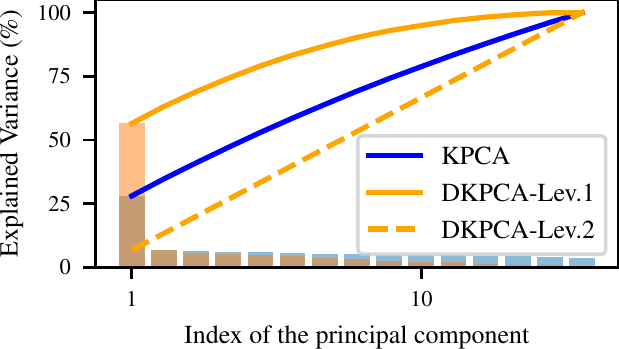}
		\caption{Synth 3.}
		\label{fig:eigs:synth3}
	\end{subfigure}
	\hfill
	\begin{subfigure}[b]{0.32\textwidth}
		\centering
		\includegraphics[width=\textwidth]{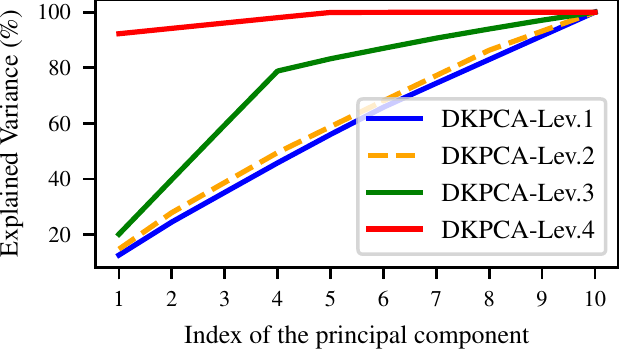}
		\caption{MNIST.}
		\label{fig:eigs:mnist}
	\end{subfigure}
	\caption{\textbf{Interpretation of the deep eigenvalues.} Explained variance (\%) of DKPCA. \textbf{(a)}  The compared method is kernel PCA with RBF kernel with the same bandwidth. Our method is able to capture considerably greater explained variance in the first principal component than shallow KPCA, showing that the proposed deeper architecture outputs more informative principal components  even with the same kernel function as the shallow KPCA. \textbf{(b)} Illustration of Lemma \ref{lemma:explainedvar:lin}: the first DKPCA level maintains higher cumulative explained variance than KPCA for all $n$, capturing more information in fewer components. \textbf{(c)} Four-level DKPCA with RBF kernels on MNIST.
In all plots, \textit{bars}: explained variance, \textit{lines}: cumulative explained variance.}
	\label{fig:eigs}
\end{figure}

\paragraph{\textbf{Deep eigenvalues}} As presented in Section \ref{sec:methods:opt}, %
deep eigenvalues $\matr{\Lambda}_{j}$ are learned by  DKPCA in  different levels $j=1, \ldots, \nlevels$, compared to  $\matr{\Lambda}$ of the single level in  shallow KPCA. 
We now investigate the learned deep eigenvalues in terms of the percentage of  variance explained and compare with shallow KPCA,
where the nonlinear case is considered by using the RBF kernel in all levels.
Fig. \ref{fig:main:explained}, \ref{fig:eigs:synth2}, and \ref{fig:eigs:synth3} plot the variance explained by each component by  DKPCA (orange bars) and by shallow KPCA (blue bars), as well as the cumulative variance explained by  DKPCA (orange line) and by shallow KPCA (blue line).

In Fig. \ref{fig:eigs:synth2} for Synth 2 with $30$ components in each level,  it shows that
the cumulative explained variance reaches almost 100\% after around 10 deep principal components, while a much slower explained variance growth in the shallow case. 
Even if both methods use the same kernels,
the first principal component of DKPCA explains around 20\% of the variance compared to only around 8\% of KPCA. This experiment shows that our method can lead to more informative principal components, ultimately resulting in a more powerful representation in fewer components with the deep architecture. Comparing the deep eigenvalues $\matr{\Lambda}_1$ (solid orange line) of the first level with the ones 
$\matr{\Lambda}_2$ (dotted orange line) of the second level,
the former shows faster initial growth, while the latter gives a flatter cumulative explained variance. %
  A similar analysis is conducted for 3DShapes in Fig. \ref{fig:main:explained}, while Fig. \ref{fig:eigs:synth3} presents the numerical evaluation of Lemma \ref{lemma:explainedvar:lin} in the two-level DKPCA with RBF first level and linear second level on Synth 3, with $\eta_2$ chosen to be the largest value satisfying the conditions of \ref{lemma:explainedvar:lin}. 
  
  Additionally, a 4-level DKPCA with 10 principal components in each level is trained on the handwritten digit images dataset MNIST \cite{lecun1998gradient} in Fig. \ref{fig:eigs:mnist}: the first and the second levels follow a similar pattern, and each subsequent level shows a flatter curve with increasingly higher explained variance in the top components. The fourth level explains almost the entire variance in the first few components, indicating that the current four levels are sufficient.
In this way, 
the minimum number of levels  to fully explain a given dataset can be determined.
This observation can also be a useful suggestion for tuning the kernel settings in the different levels: the kernel settings might need to be better tuned when introducing additional levels does not lead to a sufficient increase in explained variance.

\begin{figure}[t!]
	\centering
	\begin{subfigure}[b]{0.4\textwidth}
		\centering
		\includegraphics[scale=1]{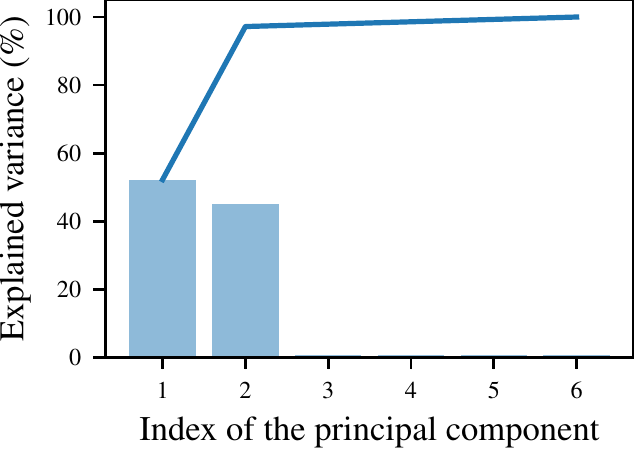}
		\caption{Explained variance.}
		\label{fig:synth1:var}
	\end{subfigure}
	\begin{subfigure}[b]{0.4\textwidth}
		\centering
		\includegraphics[scale=1]{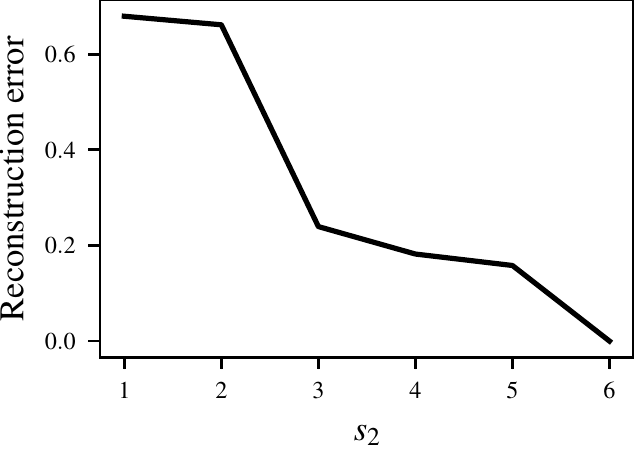}
		\caption{Reconstruction error.}
		\label{fig:synth1:recerr}
	\end{subfigure}
	\caption{\textbf{Explained variance} by the first level and \textbf{reconstruction error} (training MSE) for the Synth 1 dataset with $s_1=6$. This experiment shows the minimum number of components such that the approximation error is small enough so that practitioners have a guarantee on the faithfulness of the representation learned by the proposed  model. For this dataset, the reconstruction error is 0 with $s_1=s_2=6$. %
	}
	\label{fig:minpc}
\end{figure}

\begin{figure}[t!]
	\centering
	\includegraphics[width=0.5\textwidth]{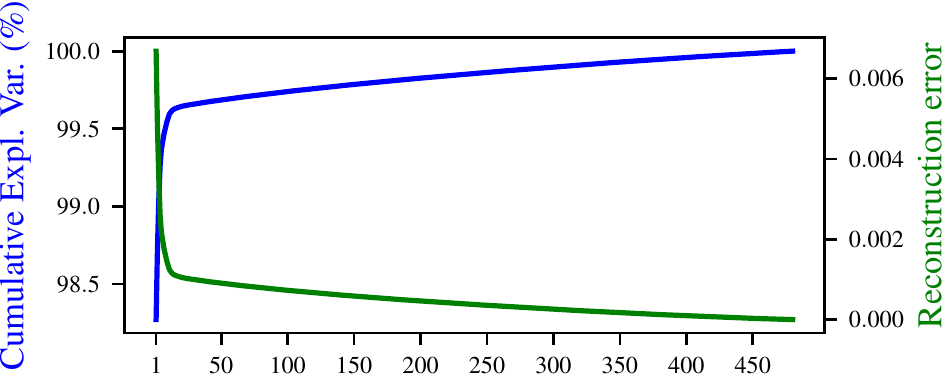} %
	\caption{\textbf{Full decomposition} for a subset of 3DShapes ($N=480$). Cumulative explained variance (\%) from the deep eigenvalues of the first level of KPCA and reconstruction error. A sharp increase in the explained variance corresponds to a distinctive drop in reconstruction error, which reaches 0 for the full decomposition. %
	}
	\label{fig:3dshapes:fullvar}
\end{figure}
\paragraph{\textbf{Selection of principal components in each level}} %
Contrary to shallow KPCA, different numbers of principal components can be selected for  each level in  deep architectures of DKPCA. In this experiment, 
we train a two-level DKPCA, introduced in Eq. \eqref{eq:system:main}, with linear kernels on the synthetic datasets to investigate the influence  of the numbers of selected principal components $s_1$ and $s_2$ of the first and second levels, respectively. In practice, one would like to select the smallest number of principal components to suffice the required small enough reconstruction error that 
depends on the specific applications, so a general method for selection of $s_j$ is needed  for practitioners. This selection can be performed by analyzing the  relative importance of each deep component through its explained variance.  

A two-dimensional synthetic dataset located as a noisy square is exemplified (Synth 1). As shown in  Fig. \ref{fig:synth1:var}, the eigenvalues of the first level drop distinctively after the second principal component, and the percentages of explained variance by the first and second component are similar. This is consistent with the ground-truth properties of this two-dimensional dataset. In Fig. \ref{fig:synth1:recerr},  the reconstruction error   decreases with $s_2$ increasing and shows its largest drop after the first two components in the second level, where the  MSE   reaches 0 with $s_1=s_2=6$. In fact, our method can always achieve 0 reconstruction error in the case of the full decomposition with  $s_1=s_2=N$, also as verified on the real-world 3DShapes in Fig. \ref{fig:3dshapes:fullvar}. For 3DShapes, the ground-truth number of variation factors is 6, so the cumulative explained variance climbs quickly as most variance has been captured by only a few components. The reconstruction error shows the opposite behavior, dropping sharply after around 10 principal components and reaching 0 for the full decomposition. 
Such evaluations are conducted in an unsupervised setting, and thus practitioners can accordingly use these evaluations to determine  $s_j$ of the DKPCA architecture for faithful reconstructions.

\paragraph{\textbf{Extracted principal components for downstream tasks}} KPCA is often used as a feature extraction step for downstream supervised tasks. Similarly, DKPCA can extract multiple levels of disentangled features that can facilitate different tasks. Specifically, it has been suggested that disentangled features could be useful for the supervised downstream problems due to the compact structure of the representation of the input distribution \cite{locatello}.
The following experiments show that DKPCA extracts more informative features that improve the performance of supervised learning problems compared to shallow KPCA. 
We fed the concatenation of the deep representation learned by an unsupervised two-level DKPCA to a linear classifier/regressor and compared with shallow (K)PCA with $s$ principal components using the same overall number of components, i.e. $s_1+s_2=s$. For all datasets $s_1=3,s_2=2$. Both KPCA and DKPCA employ RBF kernels; hyperparameters are tuned on a validation set using a 60/20/20 split for training/validation/test sets.

\begin{table*}[ht!]
    \centering
\begin{tabular}{lllll}
\toprule
Dataset & Metric & PCA & KPCA & DKPCA\\ \midrule
\multirow{2}{*}{Diabetes}     & ACC\small{($\uparrow$)}   & 70.83  & 67.89  &  \textbf{72.02} \\
                              & WINDIN\small{($\uparrow$)}& 0.001   & 0.082   &  \textbf{0.287}  \\ \midrule

\multirow{2}{*}{Ionosphere}   & ACC\small{($\uparrow$)}   & 86.09  & 92.17  &  \textbf{93.04} \\
                              & WINDIN\small{($\uparrow$)}& 0.001   & 0.22    &  \textbf{1.05}  \\ \midrule

\multirow{2}{*}{Liver}        & ACC\small{($\uparrow$)}   & 70.94  & 72.65  &  \textbf{74.36}  \\
                              & WINDIN\small{($\uparrow$)}& 0.001  & 0.069    &  \textbf{0.72}  \\ \bottomrule

\multirow{2}{*}{Cholesterol}  & RMSE\small{($\downarrow$)} & 61.67  & 61.65 & \textbf{60.59}   \\
                              & WINDIN\small{($\uparrow$)}& 0.0001   & 0.0001    &  \textbf{0.003}  \\ \bottomrule

\multirow{2}{*}{Yacht Hydrodynamics}  & RMSE\small{($\downarrow$)} & 8.41  & 8.42 & \textbf{8.02}   \\
                                      & WINDIN\small{($\uparrow$)}& 0.0001   & 0.0001    &  \textbf{0.19}  \\ \bottomrule
\end{tabular}
    \caption{Comparison of test performance for classification/regression and disentangled feature learning  by DKPCA on real-world datasets of various data types. Higher scores ({\small $\uparrow$}) are better for ACC (\%) and WINDIN, lower scores ({\small $\downarrow$}) are better for RMSE. The best performance is in bold. All datasets are UCI datasets from \cite{dua2019}.
    }
	\label{tab:sup}
\end{table*}

Results are shown in Table \ref{tab:sup}. DKPCA outperforms shallow KPCA in all datasets  in terms of both accuracy (ACC) and root mean squared error (RMSE). The WINDIN metric \cite{do2022} evaluates the disentanglement of a representation $\matr{z}$ when the ground truth factors of variations are not known: it measures both the informativeness and the separability of the representation through the  conditional mutual information between the input $\matr{x}$ and its latent representation $\matr{z}$. DKPCA produces significantly more disentangled representations than KPCA; for instance, in the Liver dataset DKPCA improves the WINDIN by approximately 10 times 
over KPCA. 
Overall, DKPCA leads to better supervised performance than KPCA while using the same number of components, showing the improved informativeness of the deep representation, which can more efficiently capture the trends of the data that are most relevant for supervised prediction.

\section{Discussion and Conclusion} \label{sec:discussion}
\subsection{Discussion}\label{sec:discussion:sub}
{
Our proposed DKPCA establishes a novel framework for  deep nonlinear principal component analysis by leveraging the RKM formulation. DKPCA exploits the Fenchel-Young inequality  introducing
conjugate feature duality, and extends the classical shallow KPCA  to  multiple levels, where both  neural network feature mappings  and  kernel functions can be adopted in different levels  for flexible modelling. 
 In contrast to shallow KPCA involving a single eigendecomposition  to the kernel matrix,  DKPCA gives different eigenvalue problems  across levels and yields the so-called  deep eigenvectors and deep eigenvalues,
 as characterized by the stationary conditions. 
 DKPCA can be applied to  general feature learning tasks in place of
classical KPCA or VAE-based methods  in various applications. Conventional KPCA may need many components to attain a high explained variance, while DKPCA can capture information more efficiently in fewer components. Compared to the black-box optimization in deep learning-based methods,  the optimization problem of DKPCA explicitly formulates
a set of nonlinear equations for each level resembling an eigenvalue problem.

DKPCA formalizes the couplings between  levels in terms of the conjugated hidden features, playing the roles of principal components  in the latent spaces with dual formulations. 
The proposed deep kernel method is not a simple forward level-wise algorithm, but the optimization of features
 flows backwards in the deep architecture, so that components in
levels of lower abstraction can benefit from the representation learned in levels of higher abstraction. This property has been theoretically verified as essential for effective hierarchical learning, and yet has not been explored in deep kernel methods. 
We then devise a multi-level optimization algorithm for DKPCA, where the deep eigenvectors and deep eigenvalues regarding the level-wise principal components  are taken as optimization variables. 
For a specific case with  two-level architectures, the optimization is simplified with solutions being
 the singular vectors in each level, 
 which facilitates theoretical analysis for greater insights: the  Eckart-Young theorem is applied to establish approximation bounds, interpreting the role of the second level  as a regularizer, and the explained variance by DKPCA is analytically compared with shallow KPCA.

We also develop the generative DKPCA, so that hidden features in multiple levels
can be sampled from the latent spaces and their correspondingly newly generated data can be attained. The role of each component or each level can be explored by traversing it in the latent space and keeping the others fixed, providing diversified aspects to explore the meaning of  principal components and the variation factors of data. The pre-image problem is a well-known challenging problem in KPCA, and its solution to general cases of multi-level KPCA was not investigated before. In DKPCA,  we incorporate the reconstruction errors, minimized to approximate the pre-image feature mappings, so that the reconstruction procedures can be conducted.  Compared to the generation and reconstruction in VAE-based methods, DKPCA creates multiple latent spaces, which not only enhances modelling flexibility with deep architectures but also provides multi-level feature learning. 
Out-of-sample extensions are also allowed in DKPCA  to predict unseen data.
 The scalability issue commonly exists in  kernel-based methods, but this can be well resolved by the out-of-sample extensions  owned by DKPCA. 
When a small subset with $M$ samples is used in training and the rest $N-M$ samples are predicted via  out-of-sample extensions, the maximal storage complexity of level $j$  drops from $\mathcal{O}(N^2)$ to  $\mathcal{O}(M^2)$.

\subsection{Conclusion}
In this paper, the proposed DKPCA introduces a novel deep architecture for unsupervised multi-level feature learning, where deep kernel machines and neural networks can both be exploited. DKPCA realizes forward and backward learning and provides more informative features enabling the exploration and interpretations on hierarchical feature abstractions. Both theoretical derivations and numerical evaluations verify the  effectiveness of DKPCA. The data representations learned by DKPCA can be utilized in various tasks and on different types of data with  promising practical values in the era of versatile data.
 In future works, variants of KPCA can be extended to deep architectures for greater efficiency or reliability, such as sparse KPCA and robust KPCA. %

\section*{Acknowledgments}
This work is jointly supported by ERC Advanced Grant E-DUALITY (787960), KU Leuven Grant CoE PFV/10/002, and Grant  FWO G0A4917N, EU H2020 ICT-48 Network TAILOR (Foundations of Trustworthy AI - Integrating Reasoning, Learning and Optimization), and Leuven.AI Institute. This work was also supported by the Research Foundation Flanders (FWO) research projects G086518N, G086318N, and G0A0920N; Fonds de la Recherche Scientifique — FNRS and the Fonds Wetenschappelijk Onderzoek — Vlaanderen under EOS Project No. 30468160 (SeLMA).

\bibliographystyle{unsrt}
\bibliography{references}

\clearpage
\appendix

\section{Proofs and Derivations} \label{sec:methods}

In this section, mathematical derivations to the modelling,  optimization, and analytical properties  of the proposed DKPCA are elaborated. DKPCA establishes a novel deep architecture of KPCA, which has long been an important unsupervised feature learning methodology.  
\ref{sec:methods:dkpca} provides the formulations leading to the optimization interpreted by a set of eigendecompositions. It demonstrates how  DKPCA leverages the RKM formulations bridging neural networks and kernels and enjoys the merits of flexible deep architectures and more interpretable kernel methods. In what follows, technical details of 
the generative modelling  are presented in 
\ref{sec:methods:gen},
showing 
promising potentials for versatile scenarios in real-world applications. Proofs for the lemmas in Section \ref{sec:results:analysis} are given in \ref{sec:methods:theory}, providing more details and insights towards  the proposed DKPCA under the considered settings with analytical properties.

\subsection{Derivation of DKPCA} \label{sec:methods:dkpca}
{ The objective \eqref{eq:deeprkm:primal:main} of DKPCA in the primal formulations is given by the compositions of latent spaces of multiple levels, and its dual formulations can be attained by characterizing the 
 stationary points to  \eqref{eq:deeprkm:primal:main}: 
\begin{equation} \label{eq:stationarity:main}
	\left\{
	\begin{array}{llll}
		\vspace{0.1cm}
		\dfrac{\partial J}{\partial \bm  h_i^{(1)}}         &= 0 &\Rightarrow& \matr{W}_1^\top  \varphi_1(\bm x_i) = \matr{\Lambda}_1 \bm h_i^{(1)} - \dfrac{\partial}{\partial \bm h_i^{(1)}} \left[ \varphi_2(\bm h_i^{(1)})^\top  \matr{W}_2\bm h_i^{(2)} \right],                                                    \\
		\vspace{0.1cm}
		\dfrac{\partial J}{\partial \matr{W}_1}               &= 0 &\Rightarrow& \matr{W}_1 = \dfrac{1}{\eta_1} \sum\limits_{i=1}^N \varphi_1(\bm x_i) {\bm h_i^{(1)}}^\top ,                            \vspace{0.1cm}                                                                                         \\
		
		\dfrac{\partial J}{\partial \bm h_i^{(j)}}         &= 0 &\Rightarrow& \matr{W}_j^\top  \varphi_j(\bm h_i^{(j)}) = \matr{\Lambda}_j \bm h_i^{(j)} - \dfrac{\partial}{\partial \bm h_i^{(j)}} \left[ \varphi_{j+1}(\bm h_i^{(j)})^\top  \matr{W}_{j+1}\bm h_i^{(j+1)} \right], \quad \forall j=2,\dots,\nlevels-1,
		\vspace{0.1cm}
		\\
		\dfrac{\partial J}{\partial \matr{W}_j}               &= 0 &\Rightarrow& \matr{W}_j = \dfrac{1}{\eta_j} \sum\limits_{i=1}^N \varphi_j(\bm h_i^{(j-1)}) {\bm h_i^{(j)}}^\top, \quad \forall j=2,\dots,\nlevels-1,                                    \vspace{0.1cm}                                                       \\
		
		\dfrac{\partial J}{\partial \bm h_i^{(\nlevels)}}  &= 0 &\Rightarrow& \matr{W}_{\nlevels}^\top  \varphi_{\nlevels}(\bm h_i^{(\nlevels-1)}) = \matr{\Lambda}_{\nlevels} \bm h_i^{(\nlevels)},                        \vspace{0.1cm}                                                                           \\
		
		\dfrac{\partial J}{\partial \matr{W}_{\nlevels}}      &= 0 &\Rightarrow& \matr{W}_{\nlevels} = \dfrac{1}{\eta_{\nlevels}} \sum\limits_{i=1}^N \varphi_{\nlevels}(\bm h_i^{(\nlevels-1)}) {\bm h_i^{(\nlevels)}}^\top .                                                                                    \\
	\end{array}
	\right.
\end{equation}

By eliminating the weight matrices $\matr{W}_j$, one obtains the following non-linear equations in the hidden features $\bm h_i^{(j)}$:
\begin{equation} \label{eq:stationarity2:main}
	\left\{
	\begin{array}{lll}
		\vspace{0.1cm}
		\text{Level 1: }  &	\dfrac{1}{\eta_1} \sum\limits_{n=1}^N \bm h_n^{(1)} k_1(\bm x_n,\bm x_i)                                                       + \dfrac{1}{\eta_2} \sum\limits_{n=1}^N \dfrac{\partial k_2(\bm h_i^{(1)}, \bm h_n^{(1)})}{\partial \bm h_i^{(1)}} {\bm h_n^{(2)}}^\top  \bm h_i^{(2)} & = \matr{\Lambda}_1 \bm h_i^{(1)},             
		\vspace{0.1cm}\\
		
		\text{Level $j$: }        &	\dfrac{1}{\eta_j} \sum\limits_{n=1}^N \bm h_n^{(j)} k_j(\bm h_n^{(j)}, \bm h_i^{(j)})  + \dfrac{1}{\eta_{j+1}} \sum\limits_{n=1}^N \dfrac{\partial k_{j+1}(\bm h_i^{(j)}, \bm h_n^{(j)})}{\partial \bm h_i^{(j)}} {\bm h_n^{(j+1)}}^\top  \bm h_i^{(j+1)}                                         & = \matr{\Lambda}_j \bm h_i^{(j)},  
		\vspace{0.1cm}\\
		
		\text{Level $\nlevels$: } &	\dfrac{1}{\eta_{\nlevels}} \sum\limits_{n=1}^N \bm h_n^{(\nlevels)} k_{\nlevels}\left(\bm h_n^{(\nlevels-1)}, \bm h_i^{(\nlevels-1)}\right) & = \matr{\Lambda}_{\nlevels} \bm h_i^{(\nlevels)},
	\end{array}
	\right.
\end{equation}
with $j=2,\dots,\nlevels-1$. 

By organizing the above \eqref{eq:stationarity2:main} into matrices, the dual formulation of DKPCA in \eqref{eq:system:nl:main} is obtained equivalently.

\subsection{Derivation of Generative DKPCA} \label{sec:methods:gen}
{For the challenging pre-image problem for multi-level nonlinear PCA,
 we propose a procedure for generative DKPCA}
from the sampled 
hidden features $\bm h^{(j)}$ in latent spaces with explicit 
feature maps: 
the feature map $\varphi_j$ of each level is known and can also be parametric with learnable parameters.

Assume that  $\varphi_j$ is invertible, with the inverse map denoted as $\varphi_j^{-1}$, and that $\bm h^{(\nlevels)}$ is given, which can be the hidden feature vector of a training or test point, or newly sampled from the latent space.
First, given the learned  $\bm h_i^{(j)}$ from the training, we introduce an additional term per level to the objective \eqref{eq:deeprkm:primal:main} for a point $\bm x$: $\frac{1}{2} \varphi_1(\bm x)^\top  \varphi_1(\bm x)$ for the first level and $\tfrac{1}{2} \varphi_j\left(\bm h^{(j-1)}\right)^\top  \varphi_j\left(\bm h^{(j-1)}\right)$ for level $j=2,\dots,\nlevels$.

Characterizing the stationary points w.r.t. $\varphi_1(
\bm x)$ and $\varphi_j\left(\bm h^{(j-1)}\right)$, we obtain
\begin{equation}
	\left\{
	\begin{array}{ll}
		\vspace{0.1cm}
		\dfrac{\partial J}{\partial \varphi_1(\bm x)} = 0         & \Rightarrow \varphi_1(\bm x) = \matr{W}_1 \bm h^{(1)},                                           \\
		
		\dfrac{\partial J}{\partial \varphi_j\left(\bm h^{(j-1)}\right)} = 0 & \Rightarrow \varphi_j\left(\bm h^{(j-1)}\right) = \matr{W}_j \bm h^{(j)}, \quad \forall j=2,\dots,\nlevels, \\
	\end{array}
	\right.
	\label{eq:oos:stationarity}
\end{equation}
{so that the feature map $\varphi_j(\cdot)$ of each level can be calculated from the given hidden features $\bm h^{(j)}$. DKPCA then generates new samples   through the inverse maps of the multiple levels, and accordingly a generated sample $\hat{\bm x}$ is attained through $\varphi_1^{-1}$ in the first level that maps $\matr W_1 \bm h^{(1)}$ back to the input space, as shown in \eqref{eq:genx:main}.

In case the inverse map $\varphi_1^{-1}$ is unknown explicitly, one can learn a pre-image map by minimization of the AutoEncoder reconstruction as described in Section \ref{sec:dkpca:gen}.

{We also developed an extension to attain the hidden features in each level corresponding to an out-of-sample point $\bm x^\star$ from the first, third, and fifth equations in \eqref{eq:stationarity:main}. For the two-level case with linear $k_2$, where it is more straightforward to obtain the out-of-sample extension, we obtain by eliminating the interconnection matrices:
$
	{\bm h^{(2)}}^\star = \frac{1}{\eta_1 \eta_2} (\matr{\Lambda}_2 - \frac{1}{\eta_2^2} \matr{H}_2^\top \matr{H}_1  \matr{\Lambda}_1^{-1} \matr{H}_1^\top \matr{H}_2)^{-1} \matr{H}_2^\top \matr{H}_1  \matr{\Lambda}_1^{-1} \matr{H}_1^\top \matr{\Phi}_1  \varphi_1(\bm x^\star)
$
and
$
	{\bm h^{(1)}}^\star = \matr{\Lambda}_1^{-1} (\frac{1}{\eta_1} \matr{H}_1^\top \matr{\Phi}_1  \varphi_1(\bm x^\star) + \frac{1}{\eta_2} \matr{H}_1^\top \matr{H}_2 {\bm h^{(2)}}^\star).
$
}

\subsection{Proof of Deep Approximation Analysis}
\label{sec:methods:theory}
In this section, we give the proofs of Lemmas \ref{lemma:lb:ub} and \ref{lemma:explainedvar:lin} in the two-level case of \eqref{eq:system:main}.

\subsubsection{Proof of approximation bounds}
\label{sec:methods:theory:bounds}
{
With the level-wise SVD interpretation to the discussed two-level cases in \eqref{eq:system:main}, the Eckart-Young theorem can be applied to both levels, deriving the approximation errors:
\begin{equation}\label{eq:proof:level1}
\left\{
\begin{array}{ll}
\norm{\matr{K}_1+\tfrac{1}{\eta_2}\matr{H}_2\matr{H}_2^\top -\matr{H}_1\matr{\Lambda}_1\matr{H}_1^\top }_F = \sqrt{\sum_{i=s_1+1}^{r_1}  {\lambda_{i}^{(1)}}^2} \\
\norm{\tfrac{1}{\eta_2}\matr{H}_1\matr{H}_1^\top -\matr{H}_2\matr{\Lambda}_2\matr{H}_2^\top }_F = \sqrt{\sum_{i=s_2+1}^{r_2}  {\lambda_{i}^{(2)}}^2}
\end{array}
\right.
\end{equation}
 with $r_1=\text{rank}(\matr{K}_1+\tfrac{1}{\eta_2}\matr{H}_2\matr{H}_2^\top )$
 and $r_2=\text{rank}(\tfrac{1}{\eta_2}\matr{H}_1\matr{H}_1^\top ).$
We fix $\eta_1=1$ and vary the regularization factor $\eta_2$. 
With orthonormality constraints in the second level, $\norm{\matr{H}_2\matr{H}_2^\top }_F=\sqrt{s_2}$,  the lower bound in Lemma  \ref{lemma:lb:ub} is obtained.}

Using the orthonormality constraints of the second level, we  {square \eqref{eq:proof:level1} and rewrite it as}
		\begin{equation}
			\begin{split}
				\norm{\matr{K}_1+\tfrac{1}{\eta_2}\matr{H}_2\matr{H}_2^\top -\matr{H}_1\matr{\Lambda}_1\matr{H}_1^\top }^2_{\fro} &= \Tr{(\matr{K}_1+\tfrac{1}{\eta_2}\matr{H}_2\matr{H}_2^\top -\matr{H}_1\matr{\Lambda}_1\matr{H}_1^\top )^2} \\
				&= \norm{\matr{K}_1-\matr{H}_1\matr{\Lambda}_1\matr{H}_1^\top }^2_{\fro} + \Tr{\left[ (\tfrac{1}{\eta_2}\matr{H}_2\matr{H}_2^\top )^2+\tfrac{2}{\eta_2}\matr{H}_2^\top \matr{K}_1\matr{H}_2-\tfrac{2}{\eta_2}\matr{H}_2^\top \matr{H}_1\matr{\Lambda}_1\matr{H}_1^\top \matr{H}_2 \right]} \\
				&= \norm{\matr{K}_1-\matr{H}_1\matr{\Lambda}_1\matr{H}_1^\top }^2_{\fro} + \frac{s_2}{\eta_2^2} + \frac{2}{\eta_2}\Tr(\matr{H}_2^\top \matr{K}_1\matr{H}_2)- \frac{2}{\eta_2}\Tr(\matr{H}_2^\top \matr{H}_1\matr{\Lambda}_1\matr{H}_1^\top \matr{H}_2),
			\end{split}
			\label{eq:frobenious_equality:main}
		\end{equation}
		with $\Tr(\matr{H}_2^\top \matr{H}_1\matr{\Lambda}_1\matr{H}_1^\top \matr{H}_2) = \norm{\sqrt{\matr{\Lambda}_1}\matr{H}_1^\top \matr{H}_2}^2_{\fro}$. By the Cauchy-Schwartz inequality, we  further obtain
		\begin{equation}
			\begin{split}
				\norm{\sqrt{\matr{\Lambda}_1}\matr{H}_1^\top \matr{H}_2}^2_{\fro} &\leq \norm{\sqrt{\matr{\Lambda}_1}}^2_{\fro} \norm{\matr{H}_1^\top }^2_{\fro} \norm{\matr{H}_2}^2_{\fro} \\
				&= \norm{\sqrt{\matr{\Lambda}_1}}^2_{\fro} \norm{\matr{H}_1}^2_{\fro} \norm{\matr{H}_2}^2_{\fro} \\
				&\leq \norm{\sqrt{\matr{\Lambda}_1}}^2_{\fro} s_1 s_2 = \left( \sum_{i=1}^{s_1} \lambda_i^{(1)} \right) s_1 s_2.
			\end{split}
			\label{eq:sqrt_lambda:main}
		\end{equation}
		Recalling that $\matr{K}_1$ is positive semi-definite, {the inequality for the upper bound when $\eta_2 > 0$ in Lemma \ref{lemma:lb:ub}} is obtained by using~\eqref{eq:sqrt_lambda:main} in~\eqref{eq:frobenious_equality:main}.
		When $\eta_2<0$ in Lemma \ref{lemma:lb:ub},  with symmetric $\matr{K}_1$, note that $\max_{\matr{H}_2^T \matr{H}_2=I} \Tr(\matr{H}_2^\top \matr{K}_1\matr{H}_2) = \sum_{i=1}^{s_2} \widetilde{\lambda}_i$, which gives the upper bound by combining with \eqref{eq:frobenious_equality:main}.
		{Therefore, the proof of deriving the bounds for the approximation analysis in Section \ref{sec:results:analysis} is completed.}

\subsubsection{Proof of explained variance lemma}
\label{sec:methods:theory:explainedvar}
{
In the two-level architecture of \eqref{eq:system:main}, let $\widetilde{\lambda}_i$ be the $i$-th largest eigenvalue of $\matr{K}_1$ and $\lambda_i$ be the $i$-th largest eigenvalue of $\matr{K}_1+\frac{1}{\eta_2}\matr{H}_2\matr{H}_2^\top$. %

In the full decomposition case ($s_1=s_2=N$), $\matr{H}_2$ is an orthogonal matrix due to the orthogonality constraints. We denote $\bm h^{(1)}_{: i}$ the $i$-th column  of $\matr{H}_1$, i.e., the eigenvector  corresponding to  $\lambda_i$. Then for each eigenvalue $\lambda_i$, we have

		\begin{equation}\label{eq:shifted:lambda}
			    \begin{split}
				        \left( \matr{K}_1 + \frac{1}{\eta_2} \matr{H}_2\matr{H}_2^\top  \right) \bm h^{(1)}_{: i} & = \lambda_i \bm h^{(1)}_{: i} \\
				        \matr{K}_1 \bm h^{(1)}_{: i} &= (\lambda_i \matr I_N - \frac{1}{\eta_2} \matr{H}_2\matr{H}_2^\top ) \bm h^{(1)}_{: i} \\
				        \matr{H}_2^\top  \matr{K}_1 \bm h^{(1)}_{: i} &= (\lambda_i \matr{H}_2^\top  - \frac{1}{\eta_2} \matr{H}_2^\top ) \bm h^{(1)}_{: i}\\
				        \matr{K}_1 \bm h^{(1)}_{: i} &= (\lambda_i - \frac{1}{\eta_2}) \bm h^{(1)}_{: i}, \\
				    \end{split}
			\end{equation}
		yielding %
		$\widetilde{\lambda}_i = \lambda_i - \frac{1}{\eta_2}$. Note that we consider $1\leq n<N$ for the cumulative explained variance in this lemma, as otherwise one would explain 100\% of the variance, resulting in  equality in \eqref{eq:var}. Further, $\lambda_i$ is constrained to be non-negative, {i.e., $\lambda_i\geq 0$,} which keeps the ratio  $\frac{\sum_{j=1}^n\lambda_j}{\sum_{i=1}^N \lambda_i}$ between 0 and 1 for a solid analysis of the explained variance. This leads to the condition 
  
		\begin{equation} \label{eq:eta2nn}
    		\eta_2 \leq -\frac{1}{\widetilde{\lambda}_N},
		\end{equation}
		where $\widetilde{\lambda}_N$ is the smallest eigenvalue of $\matr{K}_1$.

  With \eqref{eq:shifted:lambda}, the explained variance by the first $n$ deep principal components can be rewritten as
		\begin{equation} 
	\frac{\sum_{j=1}^n\lambda_j}{\sum_{i=1}^N \lambda_i} = \frac{\sum_{j=1}^n\widetilde{\lambda}_j+\frac{n}{\eta_2}}{\sum_{i=1}^N \widetilde{\lambda}_i + \frac{N}{\eta_2} },
			\end{equation}
		which is greater than the variance explained by the shallow principal components $\frac{\sum_{j=1}^n \widetilde{\lambda}_j}{\sum_{i=1}^{N} \widetilde{\lambda}_i}$ when satisfying
\begin{equation}\label{eq:ineq:analysis}
 \frac{\mya +\tfrac{n}{\eta_2}}{\myb+\tfrac{N}{\eta_2} } - \frac{\mya}{\myb} 
= \frac{n\myb-N\mya}{\eta_2\left(\myb\right)^2+N\myb} > 0.
\end{equation}

In \eqref{eq:ineq:analysis}, there exists two cases either with denominator and numerator both positive or negative. For the former case, a positive numerator gives $\tfrac{\mya}{\Tr{\matr{K}_1}}<\frac{n}{N}$. For $1\leq n<N$, the ratio between the first largest $n$ eigenvalues and the summation of all eigenvalues, i.e., $\tfrac{\mya}{\Tr{\matr{K}_1}}$, is always greater than $\frac{n}{N}$,  which is contradictory with the condition for a positive numerator.
We thereby consider the case with denominator and numerator both negative, from which one obtains the conditions $\eta_2 < -\tfrac{N}{\Tr{\matr{K}_1}}$ and $1 \leq n < \tfrac{N\mya}{\Tr{\matr{K}_1}}$, the latter of which always holds. Combining with \eqref{eq:eta2nn}, as $ -\tfrac{N}{\Tr{\matr{K}_1}} > -\tfrac{1}{\widetilde{\lambda}_N}$, the required condition of Lemma \ref{lemma:explainedvar:lin} on the explained variance remains as  \eqref{eq:eta2nn}, i.e., $\eta_2 < -\tfrac{1}{\widetilde{\lambda}_N}$.

Note that the increase of DKPCA in explained variance can now be written as
$
    \frac{n-N\frac{\sum_{j=1}^n \widetilde{\lambda}_j}{\Tr{\matr{K}_1}}}{\eta_2 \Tr{\matr{K}_1} + N}.
$
Given fixed $\eta_2, n$, and $\Tr{\matr{K}_1}$, the explained variance boost of DKPCA is more pronounced when the decay or the cumulative ratio of the first $n$ eigenvalues of $\matr{K_1}$ is not sharp,
which is often the case in complex  real-word data.
}

\section{Supplementary Empirical Evaluations} \label{sec:app}
\renewcommand{\thefigure}{B.\arabic{figure}}
\setcounter{figure}{0}
\renewcommand{\thetable}{B.\arabic{table}}
\setcounter{table}{0}
\renewcommand{\theequation}{B.\arabic{equation}}
\setcounter{equation}{0}

\subsection{Detailed Experimental Setups} \label{sec:app:expsetups}
\begin{figure}[t!]
	\centering
	\begin{subfigure}[b]{0.45\textwidth}
		\centering
		\includegraphics[width=0.8\textwidth]{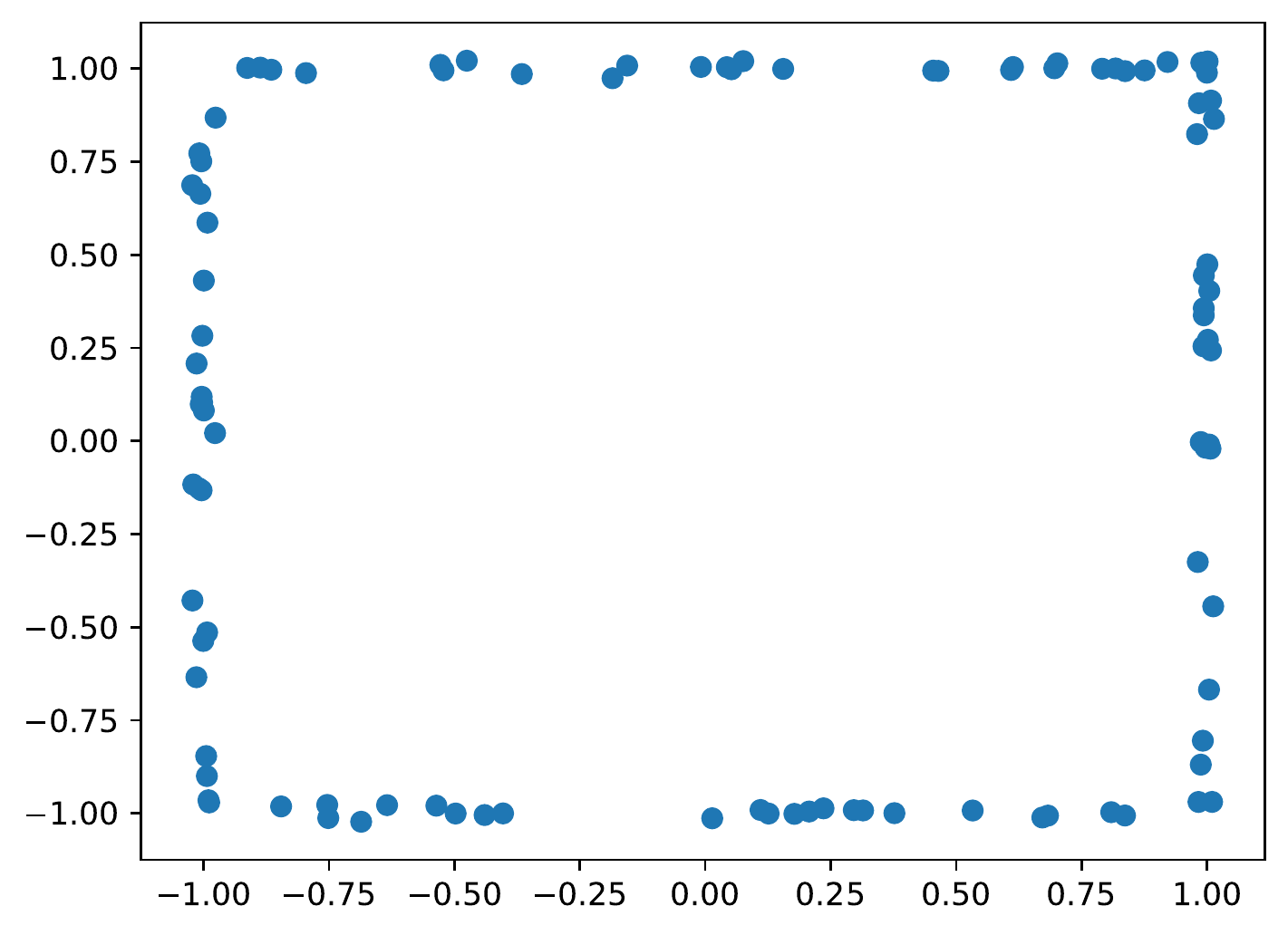}
		\caption{Square dataset (Synth 1).}
		\label{fig:square}
	\end{subfigure}
	\begin{subfigure}[b]{0.45\textwidth}
		\centering
		\includegraphics[width=.8\textwidth]{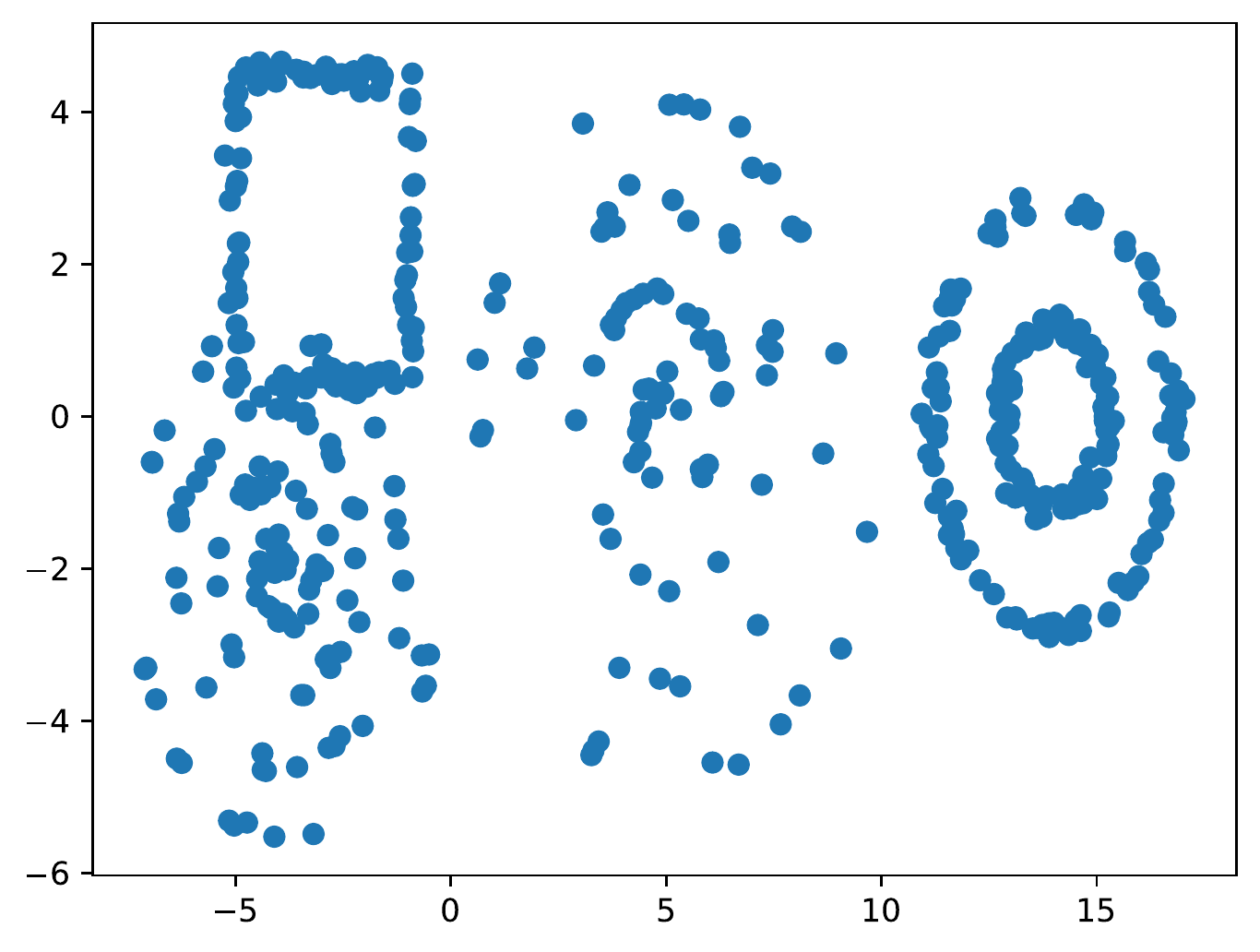}
		\caption{Synthetic dataset with spirals, ring and square (Synth 2).}
		\label{fig:complex6}
	\end{subfigure}
	\caption{Plots of the 2D synthetic data used in the experiments.}
    \label{fig:synth}
\end{figure}

\paragraph{\textbf{Datasets}} 
Three synthetic datasets are tested: a square synthetic dataset (Synth 1, see Fig. \ref{fig:square}), a complex 2D synthetic dataset consisting of one square, two spirals and one ring (Synth 2, see Fig. \ref{fig:complex6}), and a multivariate Normal synthetic dataset (Synth 3), where samples are drawn randomly from a multivariate normal distribution with zero mean and fixed covariance matrix. The test set of Synth 3 consists of $N_\text{test}=10000$ samples drawn from the same distribution.
For real-world data, 
detailed descriptions on the used datasets  can be found in Table \ref{tab:datasets}.
The downstream supervised tasks are performed on publicly available UCI benchmark datasets \cite{dua2019}.

\begin{table}[t]
\centering
	\begin{tabular}{lllll}
	\toprule
		Dataset        & Input dimensions & \begin{tabular}[c]{@{}l@{}}\# Factors of\\ variation\end{tabular} & \begin{tabular}[c]{@{}l@{}}Meaning of the factors of variation\\ and \# possible values\end{tabular}                                                                                                          & \begin{tabular}[c]{@{}l@{}}Total \# \\ data points\end{tabular} \\  \midrule
		Synth 1 & 2      & - & - & 100 \\[3pt] 
		Synth 2 & 2      & - & - & 100 \\[3pt]
		Synth 3 & 140      & - & - & 10100 \\[3pt]
		MNIST & 1 x 28 x 28      & - & - & 60000 \\[3pt]
		Cars3D         & 3 x 64 x 64      & 3                                                                 & \begin{tabular}[c]{@{}l@{}}- elevation (4 possible values)\\ - azimuth (24 possible values)\\ - object type (183 possible values)\end{tabular}                                                                & 17568 \\[20pt]
		3DShapes       & 3 x 64 x 64      & 6                                                                 & \begin{tabular}[c]{@{}l@{}}- floor color (10 possible values)\\ - wall color (10 possible values)\\ - object color (10 possible values)\\ - scale (8 possible values)\\ - shape (4 possible values)\\ - orientation (15 possible values)\end{tabular} & 480000 \\[35pt]
		SmallNORB      & 1 x 64 x 64      & 4                                                                 & \begin{tabular}[c]{@{}l@{}}- category (5 possible values)\\ - elevation (9 possible values)\\ - azimuth (18 possible values)\\ - lighting condition (6 possible values)
			 \end{tabular}                          & 4860    \\                     	 \bottomrule
			 \vspace{0.1cm}
	\end{tabular}

	\caption{Details of the datasets used in the experimental evaluation of unsupervised learning with DKPCA, where ``\#'' stands for ``number of''.}
	\label{tab:datasets}
\end{table}

\begin{table}[t!]
\centering
	\begin{tabular}{@{} l l @{}}
		\toprule
Encoder $\varphi_1(\cdot)$ & Decoder $\psi_1(\cdot)$ \\
		\midrule
		\hspace{-1mm}
	    $ \begin{cases} \makecell[l]{Conv~ [c]\times 4 \times 4; \\Conv ~[c \times 2] \times 4 \times 4;\\ Conv~ [c \times 4] \times \hat{k}\times\hat{k};     \\  FC~ 256; \\ FC~ 50 ~(Linear)}    \end{cases} $ & \hspace{-6.5mm} \quad $  \begin{cases}
		FC~256;\\
		FC~ [c \times 4] \times \hat{k}\times\hat{k};\\
		ConvTr ~[c \times 4] \times 4 \times 4;\\
		ConvTr ~[c \times 2] \times 4 \times 4;\\
		ConvTr ~[c]~ (Sigmoid)\\
		\end{cases} $ \hspace{-3mm} \\
		\bottomrule
		\vspace{0.1cm}
	\end{tabular}
\caption{Model architectures for the disentangled feature learning experiments with computer vision datasets. For all, $c=40$ and $\hat{k}=3$. All convolutions (\textit{Conv}) and transposed convolutions (\textit{ConvTr}) are with stride 2 and padding 1, except the last convolutional layer of $\varphi_1$ and the first transposed convolutional layer of $\psi_1$, which have stride 1 and no padding. Layers have Parametric-RELU ($\alpha = 0.2$) activation functions, except the output layer of the pre-image map $\psi_1$ that has Sigmoid activation function (since input data is normalized in $[0,1]$). \label{tab:archcv}}
\end{table}

\paragraph{\textbf{Evaluation metrics and compared methods}} Different related unsupervised learning methods are adopted to comprehensively evaluate our proposed deep KPCA. A comparison to the shallow kernel PCA is presented in terms of the explained variance, demonstrating the higher informativeness of the principal components learned by our method. For the general disentangled feature learning,   we consider the state-of-the-art methods $\beta$-VAE \cite{betavae}, FactorVAE \cite{factorvae}, and $\beta$-TCVAE \cite{mig}. In the qualitative disentanglement experiments, convolutional-based network architectures are used for the data feature maps, with details shown in Table \ref{tab:archcv}. We keep the same encoder $\varphi_1$ and decoder $\psi_1$ architecture for all compared methods and use $k_2(z,y)=z^\top y$ for the second DKPCA level.
For the model-specific hyperparameters, we used the suggested values in the papers of the compared methods. 
Specifically, we used $\beta=4$ for $\beta$-VAE, $\beta=6,\alpha=1,\gamma=1$ for $\beta$-TCVAE, and $\gamma=10$ for FactorVAE. 
To quantitatively evaluate the disentanglement learning, we employ the IRS metric \cite{irs}, where a higher value indicates better robustness to changes in generative factors. In other words,
if a latent variable is
associated with some generative factor, the inferred value of
this latent variable shows little change when that factor remains the same,
regardless of interventions to the other generative factors.  Other metrics for disentanglement evaluation have been proposed, but it has been shown that they are closely correlated with each other \cite{locatello}.

\paragraph{\textbf{Hyperparameter selection}} 
In unsupervised learning experiments, for consistent evaluations, the shared hyperparameters among all methods are fixed to be the same, e.g., the RBF bandwidth in KPCA. 
We fix $\eta_j=1, \, j=1,\dots,\nlevels$ and $\gamma=1$ in \eqref{eq:exp:obj:main} to equally balance the AE and deep KPCA error. For the more challenging SmallNORB, we set $\gamma=100$.
In the qualitative disentanglement experiments, we use the Riemannian Adam algorithm \cite{becigneul2019} with 80000 maximum number of epochs; concerning the principal components, $s_1=s_2$ is set to the true number of generative factors and the factors of variations involving the fewest pixels are trained on a subset with fixed factors of highest variation as these factors dominate the principal components as they have the largest number of pixels.
In the quantitative disentanglement experiments, we employ the two-level architecture of \eqref{eq:system:main} with linear kernels.
In the explained variance experiments, subsampling of 3DShapes and MNIST is performed with $N=50$ and $N=100$, respectively.
In the supervised experiments, the RBF kernel is used for all datasets. Tuning is carried out through grid search based on validation performance. The $\sigma^2$ of RBF kernels is tuned between $\exp{(-2)}$ and $\exp{(7)}$. For DKPCA, we tune $\eta_2$ between $-10$ and $10$. The hidden features of the test points are obtained through a kernel smoother approach \cite{hastie2009} for the supervised and the large-scale disentanglement experiments. The shared hyperparameters in the compared methods are tuned under the same settings, e.g., the kernel parameters are tuned in the same range.

\subsection{Additional Results} \label{sec:app:exp}
Table \ref{tab:table1} gives the test reconstruction errors on {a 140-dimensional synthetic dataset }
with different numbers of principal components in the two-level DKPCA of Eq. \eqref{eq:system:main} with linear kernels. In Table \ref{tab:table1}, for a fixed $s_1$, the best test error is obtained with $s_2 = s_1$: the test error does not further decrease for $s_2 > s_1$. In fact, 
the rank of $\matr{H}_1\matr{H}_1^T$ is at most $s_1$.
so an $\matr{H}_2$ with rank higher than $s_1$ cannot lead to lower approximation error. Therefore, for a fixed $s_1$ in {this conceived two-level architecture with linear kernels,}
$s_1$ should be set as $s_1 \leq s_2$
in terms of reconstruction error, in which increasing $s_1$ leads to lower reconstruction error as more principal components are incorporated.

\begin{table}[t!]
	\centering
		\begin{tabular}{l|llllll}
		\toprule
			\backslashbox{$s_2$}{$s_1$} & 2    & 4    & 16   & 32   & 64   & 100  \\ \hline
			2                           & 1.59 & 1.60 & 1.59 & 1.63 & 1.65 & 1.66 \\
			4                           & 1.59 & 1.53 & 1.55 & 1.62 & 1.62 & 1.64 \\
			16                          & 1.59 & 1.53 & 1.26 & 1.38 & 1.55 & 1.59 \\
			32                          & 1.59 & 1.53 & 1.26 & 1.08 & 1.35 & 1.44 \\
			64                          & 1.59 & 1.53 & 1.26 & 1.08 & 0.98 & 1.19 \\
			100                         & 1.59 & 1.53 & 1.26 & 1.08 & 0.98 & 0.97\\
			\bottomrule
		\end{tabular}
		\vspace{0.2cm}
	\caption{Test reconstruction error (MSE) on the 140D Synth 3 dataset for different numbers of principal components of the two levels in the proposed deep KPCA. All numbers are $\times 10^2$.}
	\label{tab:table1}
\end{table}

\end{document}